\documentclass[runningheads]{llncs}

\usepackage[final,year=2026,ID=458]{accv}

\usepackage{accvabbrv}
\usepackage{graphicx}
\usepackage{booktabs}
\usepackage{multirow}
\usepackage{tabularx}
\usepackage{amsmath,amssymb}
\usepackage{mathtools}
\usepackage{bm}
\usepackage[accsupp]{axessibility}
\usepackage{subcaption}
\usepackage{xcolor}
\usepackage[breaklinks,colorlinks,citecolor=accvblue]{hyperref}
\usepackage{orcidlink}

\newcommand{\Ln}{\mathbb{L}^{n}_{\kappa}}

\newcommand{\linner}[2]{\langle #1, #2 \rangle_{\mathcal{L}}}

\newcommand{\IoA}{\mathrm{IoA}}

\begin{document}

\title{The Gate Always Closes: On Injecting Auxiliary Signals into Frozen Vision-Language Models}
\titlerunning{The Gate Always Closes}

\author{Moshiur Farazi\inst{1}\orcidlink{0000-0003-1494-5921} \and
Sameera Ramasinghe\inst{2}\orcidlink{0000-0002-3200-9291} \and
Bekir Sait Ciftler\inst{1}\orcidlink{0000-0003-0229-0745} \and \\
Mahbub Ahmed Turza\inst{3}\orcidlink{0009-0005-1749-1453} \and
Shafin Rahman\inst{3}\orcidlink{0000-0001-7169-0318}}
\authorrunning{M.~Farazi et al.}
\institute{University of Doha for Science and Technology, Doha, Qatar \and
Pluralis Research, Australia \and
North South University, Dhaka, Bangladesh\\
\texttt{moshiur.farazi@udst.edu.qa}}

\maketitle

\begin{abstract}
Auxiliary signal pathways in VLMs are routinely fitted with learnable gates so the optimiser can decide how much of the signal to admit. We study what the optimiser decides, and whether the model then uses the signal, with hyperbolic visual relational graphs as the experimental vehicle. We find that under cross-entropy fine-tuning the optimiser decides on zero: the gate closes in every design we trained, across five injection designs on LLaVA-1.5-7B, 15 of 15 runs on Qwen2-VL-7B spanning four task families, 3 of 3 seeds on InternVL3-8B, and with LoRA on all linear layers. We attribute this \textbf{suppression phenomenon} to two regimes, a \emph{dead-gradient} regime formalised through the caption-invariance of image-derived signals and measured directly, and a \emph{negative-utility} regime in which the auxiliary signal actively hurts the loss. A closed gate does not by itself mean an inert pathway, so we ablate each pathway at inference under one protocol with real region proposals: the norm-aligned designs are \emph{behaviourally closed} (at most 1.5pp change at the learned gate), whereas the un-normalised gated prefix is open and harmful, costing 6.7pp on relational GQA at its learned gate of $+0.005$ while helping SugarCrepe. Evaluating every configuration under one inference mode shows that an active injected pathway costs up to 15pp on relational GQA and 7--12pp on out-of-distribution VSR, that the geometric training losses give no measurable in-distribution gain once the pathway is dropped, and that a Euclidean containment control ties the hyperbolic one. Embedding-norm alignment remains necessary for prefix injection that preserves generation. We recommend testing signal use by ablation rather than by reading gate values or attention mass. We open-source the implementation, the checkpoints and the result files behind every table at \url{https://github.com/mrfarazi/the-gate-always-closes}.
\keywords{Vision-Language Models \and Auxiliary Signal Injection \and Learnable Gates \and LoRA \and Compositional Reasoning}
\end{abstract}

\section{Introduction}
\label{sec:intro}

Parameter-efficient fine-tuning of Vision-Language Models with LoRA~\cite{liu2024llava15} is now standard practice, and auxiliary signals (scene graphs, relational features, geometric priors) are routinely injected into the frozen language model through a pathway fitted with a learnable gate, so that the optimiser can decide how much of the signal to admit~\cite{gao2023llamaadapter,wang2024llavassg,herzig2023structured,hypervis2026}. This paper asks two questions that are usually conflated: what does the optimiser decide, and does the model then \emph{use} the signal?

We use the hyperbolic relational graph of~\cite{hypervis2026} as the auxiliary signal. Region proposals are related on the Lorentz hyperboloid, IoA-driven entailment cones and angular repulsion shape the geometry, and the result reaches the LLM either as an additive bias on image-patch self-attention or as gated prefix tokens (Sec.~\ref{sec:method}). The manifold is an experimental vehicle rather than the contribution: a control that optimises the same containment objective in flat space matches it (Sec.~\ref{sec:geometry}). Our findings are four.

\noindent\textbf{The gate closes under cross-entropy.} In every configuration we trained, the learnable scalar on the pathway ends at or near zero: five injection designs on LLaVA-1.5-7B (attention biases, gated prefixes, a margin-trained objective), 15 of 15 runs of the gated-prefix design on Qwen2-VL-7B across four task families, 3 of 3 seeds on InternVL3-8B, and LoRA on all seven linear projections as a $4.8\times$ capacity control (Sec.~\ref{sec:suppression}). We call this the \textbf{suppression phenomenon} and attribute it to a first-order \emph{dead-gradient} regime, which caption-invariance of an image-derived signal makes exact for pairwise objectives and which we measure, and to a second-order \emph{negative-utility} regime in which the signal hurts the loss. The title names this regularity as we observed it: a statement about the gate parameter, under cross-entropy, in the designs and backbones we trained, which the next finding shows is not the whole story.

\noindent\textbf{Closed is not the same as unused, and neither is visible from the parameter.} A ``closed'' softplus or tanh gate can still nominally pass 30--45\% of the signal, so we ablate each pathway at inference with real region proposals and one implementation for both arms (Sec.~\ref{sec:knockout}). The norm-aligned designs are behaviourally closed (at most 1.5pp change at the learned gate). The un-normalised gated prefix is not: at its learned gate of $+0.005$ it costs 6.7pp on relational GQA and helps SugarCrepe, and forcing it fully open collapses generation. Attention onto the injected slots is ${\sim}40\times$ chance whether or not they carry content, so attention cannot indicate use either.

\noindent\textbf{Injecting at inference hurts the reasoning the signal targets.} Under one inference mode, dropping the trained pathway recovers up to 15pp on relational GQA and 7--12pp on out-of-distribution VSR for every prefix design; the geometric training losses give no measurable in-distribution gain once the pathway is dropped, and flat-space containment ties the hyperbolic manifold (Sec.~\ref{sec:results}).

\noindent\textbf{Norm alignment is necessary for prefix injection that preserves generation.} Raw hyperbolic prefixes carry 36$\times$ the text-embedding norm, dominate the residual stream and collapse generative VQA; RMS alignment to the text scale removes the collapse (Sec.~\ref{sec:norm_fix}).

\noindent\textbf{Positioning.} The closest prior threads are prefix and prompt tuning~\cite{jia2022vpt,gao2023llamaadapter}, auxiliary signal injection into VLMs via scene graphs, prefix tokens or structural attention biases~\cite{wang2024llavassg,mitra2024ccot,herzig2023structured,hypervis2026,yang2026scenegraph_thinking}, hyperbolic VLMs and entailment-cone geometry~\cite{desai2023meru,pal2025hycoclip,ramasinghe2024accept,peng2025hyperet}, structured attention biases in the ALiBi/RoPE family~\cite{press2022alibi,su2024rope,wu2025flashbias,dalal2026attwarp}, compositionality benchmarks~\cite{yuksekgonul2023aro,thrush2022winoground,hsieh2024sugarcrepe,dumpala2024sugarcrepepp,mishra2025scramble}, and noise-as-regularisation~\cite{jain2024neftune}. All of these treat a shaped auxiliary signal as something to be consumed at inference through a (usually gated) pathway. We differ on two fronts: we characterise the conditions under which the optimiser instead closes that pathway, and we measure whether the model uses whatever signal still gets through. The full literature review is in the supplementary material.

\section{Experimental Vehicle: Hyperbolic Relational Encoding}
\label{sec:method}

\begin{figure*}[t]
    \centering
    \parbox[c][6cm][c]{\linewidth}{\centering\includegraphics[width=\linewidth,height=6cm,keepaspectratio]{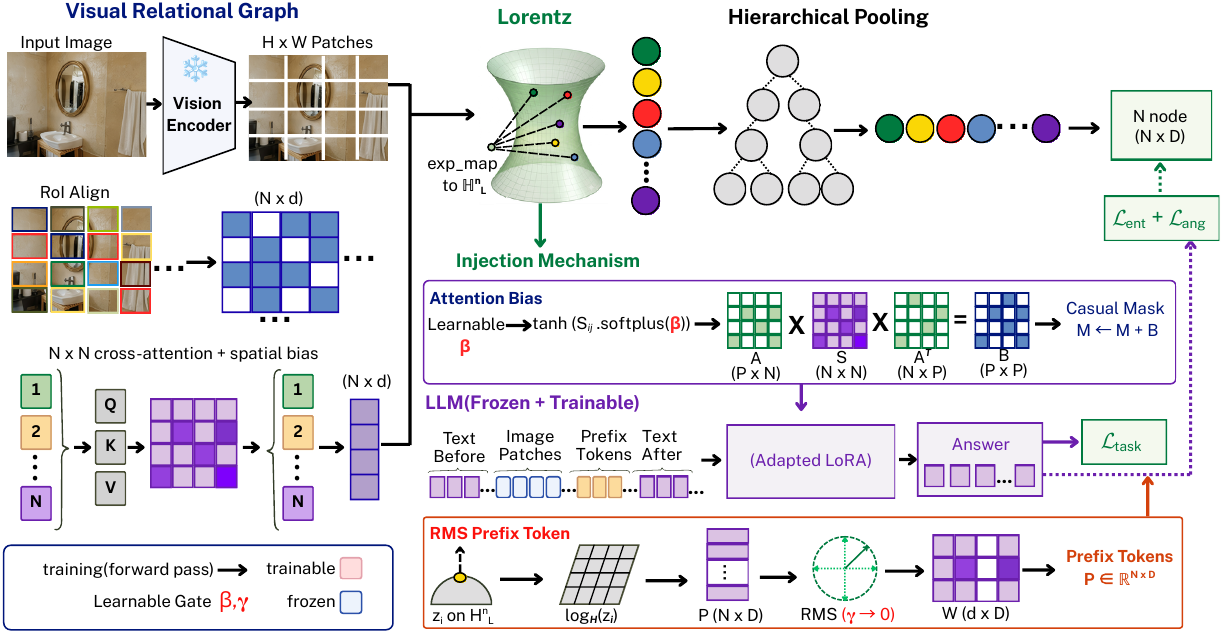}}
    \caption{Pipeline overview. From class-agnostic region proposals, we compute dense visual relations, embed them on the Lorentz hyperboloid, and route the geometric signal through two injection mechanisms: attention bias (Mechanism A) and gated prefix tokens (Mechanism B). Auxiliary losses $\mathcal{L}_\text{ent}$ and $\mathcal{L}_\text{ang}$ act on the graph during training; both gates $\beta,\gamma$ close under the task loss.}
    \label{fig:architecture}
\end{figure*}

We use a hyperbolic relational graph~\cite{hypervis2026} as the source of the auxiliary signal. The architecture (Fig.~\ref{fig:architecture}) is an experimental vehicle rather than the contribution.

\subsection{Manifold and Relational Graph}

We work in the Lorentz model $\Ln$, the upper sheet of the $(n{+}1)$-dimensional hyperboloid ($n{=}256$) with learnable curvature $\kappa>0$ (initialised at 1.0, clamped to $[0.1, 5.0]$). The exponential map at the origin $\mathbf{o}$ sends tangent vectors $\mathbf{v}$ (with $v_0{=}0$) to manifold points; every point $\mathbf{q} \in \Ln$ with $\|\tilde{\mathbf{q}}\| \ge K{=}0.1$ defines an entailment cone~\cite{ganea2018hyperbolicnn,ramasinghe2024accept} with half-aperture $\omega(\mathbf{q}) = \arcsin(K / \|\tilde{\mathbf{q}}\|)$, and two points are separated by the central angle $\phi(\mathbf{p},\mathbf{q})$ at the origin (Euclidean angle between the spatial parts of $\log_\mathbf{o}^\kappa(\mathbf{p})$ and $\log_\mathbf{o}^\kappa(\mathbf{q})$). The cone containment test $\phi(\mathbf{p},\mathbf{q}) \le \omega(\mathbf{q})$ is inherently asymmetric.

From $N{=}36$ class-agnostic region proposals we extract RoI features $\mathbf{v}_i \in \mathbb{R}^{d_v}$ ($d_v{=}1024$, CLIP ViT-L, before the multi-modal projector) and compute pairwise relation features by spatially-biased multi-head self-attention over relative geometry (centre offsets, log size ratios), giving a dense relation tensor $\mathbf{R} \in \mathbb{R}^{N \times N \times d}$ ($d{=}256$). Each $\mathbf{r}_{ij}$ is projected to a tangent vector (norm clamped to 5), mapped to $\Ln$ by the exponential map at the origin, and aggregated per node through the Einstein midpoint~\cite{ungar2005analytic}. Full definitions are in the supplementary material.

\subsection{Geometric Losses}
\label{sec:losses}

Following Ramasinghe \etal~\cite{ramasinghe2024accept}, we use purely geometric losses that operate on angles rather than geodesic distances, since the latter conflict with entailment and drive the curvature toward collapse. We denote the asymmetric Intersection-over-Area of region $a$ into region $b$ as $\IoA(a {\to} b) := |a \cap b| \,/\, |a|$. When $\IoA(a {\to} b) > 0.8$, region $a$ is spatially contained in $b$ and its embedding should fall inside $b$'s entailment cone:
\begin{equation}
\mathcal{L}_{\text{ent}} = \mathbb{E}_{(a,b): \IoA > 0.8} \max\!\big(0,\; \phi(\mathbf{z}_a, \mathbf{z}_b) - \eta\,\omega(\mathbf{z}_b)\big).
\label{eq:ent}
\end{equation}
If instead $\IoA < 0.05$ in both directions, the regions are disjoint and we push their embeddings apart in angle:
\begin{equation}
\mathcal{L}_{\text{ang}} = \mathbb{E}_{(a,b): \IoA < 0.05} \max\!\big(0,\; \tfrac{\pi}{2} - \phi(\mathbf{z}_a, \mathbf{z}_b)\big).
\label{eq:ang}
\end{equation}
A hinge penalty $\mathcal{L}_\kappa = \max(0, 0.3 - |\kappa|)^2$ prevents curvature collapse. The full objective is $\mathcal{L} = \mathcal{L}_{\text{task}} + \alpha(t)(\lambda_{\text{ang}}\mathcal{L}_{\text{ang}} + \lambda_{\text{ent}}\mathcal{L}_{\text{ent}} + \lambda_\kappa \mathcal{L}_\kappa)$, where $\alpha(t)$ ramps linearly from 0.001 to $\alpha_{\max}$ over 100 steps.

\subsection{Two Injection Mechanisms}
\label{sec:mechanisms}

We test two ways to deliver the geometric signal to the LLM. Both are formalised here in the notation that the gradient analysis of Sec.~\ref{sec:suppression} requires.

\noindent\textbf{Mechanism A: Attention bias.} From the node embeddings $\{\mathbf{z}_i\}_{i=1}^N$ we compute asymmetric cone scores
\begin{equation}
s_{ij} = \eta \cdot \omega(\mathbf{z}_j) - \phi(\mathbf{z}_i, \mathbf{z}_j), \qquad \hat{s}_{ij} = \tanh\!\big(s_{ij} \cdot \mathrm{softplus}(\beta)\big),
\label{eq:cone}
\end{equation}
where $\eta$ (init 1.5) and $\beta$ (init 0) are learnable. The scores are masked to valid proposal pairs and mapped to a $P{\times}P$ patch-level bias ($P{=}576$) through a soft IoU-based assignment $\mathbf{A} \in \mathbb{R}^{P \times N}$ and bilinear interpolation, $\mathbf{B} = \mathbf{A}\hat{\mathbf{S}}\mathbf{A}^\top$. During prefill, this bias is added to the LLM's causal mask at image-patch positions. The property that matters for the suppression analysis is simple: $\mathbf{B}$ depends only on the image, not on the text.

\noindent\textbf{Mechanism B: Gated prefix tokens.} A hyperbolic Top-$K$ gate selects the $K{=}4$ relation embeddings closest in geodesic distance to a learnable query $\mathbf{q} \in \Ln$, projects them to LLM dimension via $\mathbf{W}_p \in \mathbb{R}^{d \times d_{\text{llm}}}$, applies RMS normalization, and scales by $\tanh(\gamma)$ with $\gamma$ (init 0) a learnable gate:
\begin{equation}
\mathbf{t}_k = \tanh(\gamma) \cdot \mathrm{RMSNorm}\!\big(\mathbf{W}_p \tilde{\mathbf{u}}_k\big) \cdot \frac{\|\bar{\mathbf{e}}\|_2}{\|\mathrm{RMSNorm}(\mathbf{W}_p \tilde{\mathbf{u}}_k)\|_2},
\label{eq:prefix}
\end{equation}
where $\tilde{\mathbf{u}}_k = (\log_\mathbf{o}^\kappa \mathbf{z}_{i_k})_{1:n}$ is the spatial part of the tangent vector of the $k$-th selected embedding and $\bar{\mathbf{e}}$ is the mean text embedding. The resulting tokens are inserted after the image patches. During training, prefix dropout ($p{=}0.5$) randomly zeros the prefix tokens.

\noindent\textbf{Training setup.} LLaVA-1.5-7B backbone with LoRA (rank 16, \texttt{q\_proj}/\texttt{v\_proj}). Three optimizer groups: AdamW for LoRA ($2{\times}10^{-5}$), Riemannian Adam~\cite{geoopt} for the graph ($10^{-4}$), and vanilla Adam for $\kappa$ ($10^{-4}$, no gradient clipping). Training runs on GQA \texttt{train\_balanced} (943{,}000 questions over 72{,}140 images; scene-graph boxes are available for every training image) for 3--5 epochs. Compute details are in the supplementary material.

\noindent\textbf{Controls.} We also train: a \emph{soft prompt} (same gated-prefix architecture, randomly initialised learned embeddings, no geometric backbone, matched parameter count); a \emph{Euclidean gated} prefix (the relational graph in flat space, arithmetic means for Einstein midpoints, trained \emph{without} geometric losses); a \emph{Euclidean containment} control (the flat-space graph trained with $\mathcal{L}_{\text{ent}}$ and $\mathcal{L}_{\text{ang}}$ on cosine angles, three seeds); the hyperbolic prefix with the geometric losses removed (\emph{G2}); an RMS-normalised \emph{gated} prefix completing the $2{\times}2$ of Sec.~\ref{sec:norm_fix}; and a LoRA-on-all-linear-layers variant (Sec.~\ref{sec:backbones}).

\section{The Suppression Phenomenon}
\label{sec:suppression}

In every configuration we trained, a learnable scalar controlling an auxiliary, caption-invariant pathway ended near zero under the task loss. We state the protocol (Sec.~\ref{sec:protocol}), document closure on three backbones (Secs.~\ref{sec:designs}--\ref{sec:backbones}), give the mechanism (Sec.~\ref{sec:regimes}), and then ask separately whether a closed pathway is \emph{unused} (Sec.~\ref{sec:knockout}). The parameter alone cannot answer that: $\mathrm{softplus}(-0.60)$ still passes ${\sim}44\%$ of the signal.

\subsection{Evaluation Protocol}
\label{sec:protocol}

Three conventions hold for every behavioural number in this paper.

\noindent\textbf{Proposals.} GQA testdev, SugarCrepe (COCO), VSR and POPE carry no scene-graph annotations, so every behavioural comparison uses a fixed $6{\times}6$ grid of region proposals, passed through a single implementation for both arms; on the S1 checkpoint the resulting bias has mean magnitude 0.4 and maximum 0.8 logits. Training used real scene-graph boxes throughout, so gate values and every LoRA-only number are independent of this choice.

\noindent\textbf{Inference mode.} A configuration is evaluated with the pathway \emph{on} (bias at its learned scale; prefix at its learned gate, or at scale~1 for ungated prefixes), \emph{zeroed} (bias ${\times}0$; prefix tokens multiplied by zero with their slots kept), or \emph{absent} (no bias; no prefix slots, i.e.\ LoRA-only). Every table states which. Keeping zeroed slots costs 2--5pp of relational GQA relative to absent slots for every prefix design trained with LoRA on the query and value projections (see the supplementary material), so \emph{zeroed} is the reference for whether the content is used and \emph{absent} for whether the pathway should exist.

\noindent\textbf{Statistics.} Paired bootstrap on shared items (5{,}000 resamples for the $n{=}1{,}000$ ablations, 10{,}000 for the matched tables), 95\% intervals in brackets; every table states its evaluation set and $n$. ``GQA-rel'' means testdev questions of semantic type \texttt{rel}.

\subsection{Five Demonstrations on LLaVA-1.5-7B}
\label{sec:designs}

\noindent\textbf{(S1) Attention bias, end-to-end ($\alpha_{\max}{=}0.05$):} $\beta$ is initialised at 0 and trained jointly with the task and geometric losses. Over 5 epochs it settles at $-0.70$ ($\mathrm{softplus}(-0.70){\approx}0.41$).

\noindent\textbf{(S2) Attention bias, reduced geometric weight ($\alpha_{\max}{=}0.01$):} $\beta$ settles at $-0.60$. Full-testdev GQA with the bias absent is 60.16\%, at the vanilla level (Table~\ref{tab:main}), so the interference of the geometric gradient with LoRA, not the bias itself, was what cost S1 its 2pp.

\noindent\textbf{(S3) Compositional margin loss (Stage 2):} LoRA frozen, only $\beta$ trainable, margin loss on positive/negative caption pairs. $\beta$ declines monotonically from 0 to $-8.50$ over 50K steps ($\mathrm{softplus}(-8.50) \approx 2{\times}10^{-4}$). Here, and only here, the parametric and behavioural senses of ``closed'' coincide by construction.

\noindent\textbf{(S4) Gated prefix, no dropout:} prefix tokens derived from the hyperbolic graph, RMS-normalised, gated by $\tanh(\gamma)$; $\gamma$ settles at $-0.38$ ($\tanh\gamma{\approx}{-}0.36$), an inverted low-magnitude signal.

\noindent\textbf{(S5) Raw (unnormalised) prefix, gated:} the same tokens without RMS rescaling, gated by $\tanh(\gamma)$ with $\gamma$ initialised at 0. $\gamma$ never moves off zero (final $+0.005$, Fig.~\ref{fig:suppression}); this is the stalling regime formalised as Regime~A below.

\noindent Three further gated runs behave the same way: the RMS-normalised gated prefix with dropout ends at $\gamma{=}{-}0.15$, and two further soft-prompt runs at $-0.26$ and $+0.12$ (first run $-0.20$).

\subsection{Two Further Backbones and a Capacity Control}
\label{sec:backbones}

\begin{table}[t]
\centering
\caption{Gate closure across backbones and adapter capacity. ``Forced to'' sets the gate of the closed checkpoint to the stated value; effects are against the zeroed pathway, paired 95\% CIs. LLaVA and Qwen: GQA-rel ($n{=}1{,}000$ / $5{,}308$); InternVL: teacher-forced cross-entropy, $n{=}300$, $\Delta$CE $>0$ = worse. Qwen2-VL-7B and InternVL3-8B use the S5 recipe. $^{a}$S4, the RMS-gated base run, two soft prompts, the Euclidean gated prefix, two LoRA-all-linear runs.}
\label{tab:backbones}
\vspace{-0.8em}
{\footnotesize\setlength{\tabcolsep}{3pt}
\begin{tabularx}{\columnwidth}{@{}>{\raggedright\arraybackslash}p{1.35cm} *{5}{>{\raggedright\arraybackslash}X}@{}}
\toprule
 & \multicolumn{3}{c}{LLaVA-1.5-7B} & \mbox{Qwen2-VL-7B} & \mbox{InternVL3-8B} \\
\cmidrule(lr){2-4}\cmidrule(lr){5-5}\cmidrule(lr){6-6}
Design & attention bias (S1 / S2 / S3) & RMS-gated prefix$^{a}$ & raw gated prefix (S5) & S5 recipe (5 tasks $\times$ 3 seeds) & S5 recipe (3 seeds) \\
\midrule
Runs & 3 & 7 & 1 & 15 & 3 \\
Final gate(s) & $\beta$: $-0.696$ /\newline $-0.600$ /\newline $-8.496$ & $\tanh\gamma$: $-0.362$ to $+0.185$ & $\tanh\gamma$: $+0.005$ & all $|\tanh\gamma|\le0.018$ & $\tanh\gamma$: $+0.006$ /\newline $-0.001$ /\newline $-0.004$ \\
Forced to & bias $\times3$ & $\gamma{=}1$ & learned /\newline $\gamma{=}1$ & $\gamma{=}1.5$ /\newline $\gamma{=}2.0$ & $\gamma{=}2$ \\
Effect [95\% CI] & $+1.0$ $[-0.1, 2.1]$\newline $+0.1$ $[-0.8, 1.0]$\newline $+0.0$ $[-0.5, 0.5]$ & $-0.7$ to $+3.4$;\newline learned gate $|\Delta|\le1.5$ & $-6.7$ $[-8.4, -5.0]$ /\newline $-38.5$ $[-41.7, -35.2]$ & $+0.17$ $[-0.21, 0.53]$ /\newline $+0.15$ $[-0.24, 0.55]$ & $\Delta$CE: $-0.007$ $[-0.026, 0.009]$ /\newline $+0.014$ $[0.001, 0.030]$ /\newline $+0.001$ $[-0.011, 0.013]$ \\
\bottomrule
\end{tabularx}
}
\end{table}

Is closure a property of one backbone, or of a small adapter? We ported the S5 recipe to \textbf{Qwen2-VL-7B}~\cite{wang2024qwen2vl}, which differs from LLaVA-1.5 in pre-norm layout, vision encoder (dynamic patching with patch merging), positional scheme (3D M-RoPE) and pretraining scale, and to \textbf{InternVL3-8B}~\cite{zhu2025internvl3}. On Qwen the gate closes in 15 of 15 runs across four task families and under a de-trapped additive reparameterisation that removes the RMSNorm well of Sec.~\ref{sec:regimes}, so closure is not a normalisation artefact; forcing $\gamma{=}1.5$ or $2.0$ on a closed checkpoint moves GQA-rel by $+0.17$pp $[-0.21,+0.53]$ and $+0.15$pp $[-0.24,+0.55]$ ($n{=}5{,}308$). On InternVL3-8B the gate closes in 3 of 3 seeds ($|\tanh\gamma|\le0.006$), and forcing it open never helps: inert in two seeds, a small significant degradation in the third. LoRA on all seven linear projections of LLaVA ($39.98$M vs.\ $8.39$M parameters) leaves the gate closed in both seeds ($+0.024$, $-0.030$), drifting in opposite directions: what replicates is closure, not a sign, and S3, with LoRA frozen, already excludes capacity. Closure is \emph{not} expected for text-coupled pathways whose contribution depends on the caption, as in LLaMA-Adapter~\cite{gao2023llamaadapter}. Only the S5 recipe was ported.

\subsection{First- vs.\ Second-Order Suppression Regimes}
\label{sec:regimes}

\begin{figure}[t]
    \centering
    \parbox[c][5cm][c]{\linewidth}{\centering\includegraphics[width=\linewidth,height=5cm,keepaspectratio]{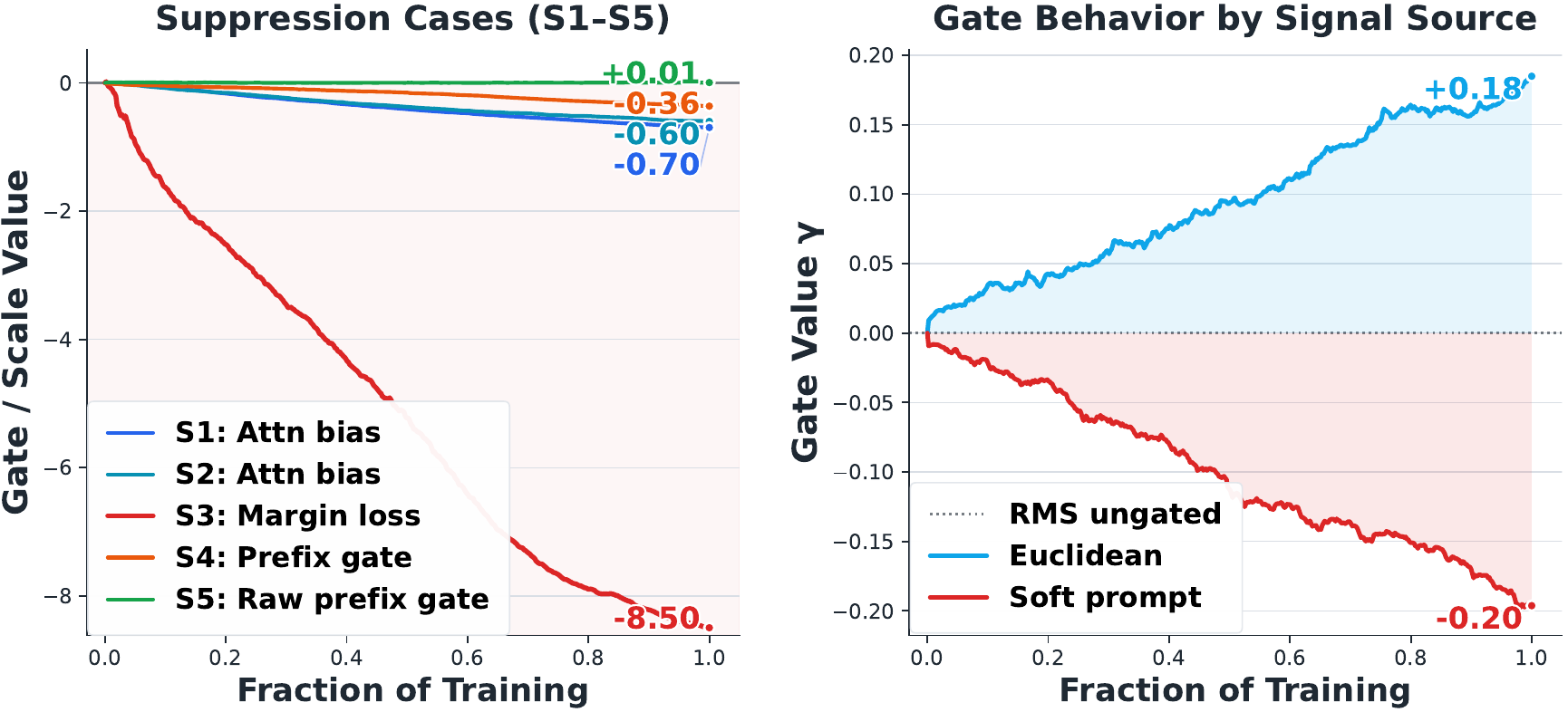}}
    \vspace{-0.4em}
    \caption{Gate trajectories during training. \textbf{Left:} the five designs of Sec.~\ref{sec:designs}; S1, S2 and S4 drift to modest negatives, S3 dives monotonically to $\beta \to -8.5$, S5 stalls near zero. \textbf{Right:} the two gated controls; the Euclidean-gated prefix is the only gate that ends positive ($\gamma{=}{+}0.19$), the soft-prompt gate ends at $-0.20$. Whether an endpoint is also \emph{behaviourally} closed is a separate question, answered in Table~\ref{tab:ablations_main}.}
    \label{fig:suppression}
\end{figure}

The designs exhibit qualitatively different dynamics (Fig.~\ref{fig:suppression}): S5 stalls near zero throughout training; S3 dives monotonically to $-8.5$; S1, S2 and S4 drift to modest negatives. Proposition~\ref{prop:suppression} isolates the first-order discriminative gradient through the gate; the step from a vanishing first-order term to a fixed point requires an assumption about higher-order terms that S3 falsifies, so two regimes must be distinguished.

\noindent\textbf{Notation.} Let $f_\theta$ denote the LLM and let $\mathbf{b}(I) \in \mathbb{R}^m$ be an image-derived auxiliary signal entering the LLM through a scalar gate $\gamma$. We use the additive linearisation $\mathbf{h}(I, c; \gamma) \approx \mathbf{h}_0(I, c) + \gamma \cdot g(\mathbf{b}(I))$ as an idealisation of both injection paths; higher-order corrections in $\gamma$ mediated by softmax and residual nonlinearities are what distinguish the two regimes. Let $\ell$ be the sequence log-likelihood readout and $\mathcal{L}_{\mathrm{pair}} = \max(0, m - (s^+ - s^-))$ with $s^\pm = \ell(\mathbf{h}(I, c^\pm; \gamma))$.

\begin{lemma}[Exact factorisation under caption-invariance]
\label{lem:factor}
If $\partial \mathbf{b}/\partial c = 0$, then
$\partial(s^+ - s^-)/\partial \gamma\,|_{\gamma=0} = \langle \Delta(I),\, g(\mathbf{b}(I)) \rangle$ with $\Delta(I) := \partial\ell/\partial\mathbf{h}|_{c^+} - \partial\ell/\partial\mathbf{h}|_{c^-}$.
\end{lemma}

\noindent Caption-invariance does exactly one thing: it makes the injected vector the \emph{same} $g(\mathbf{b}(I))$ in both branches, so it factors out of the difference; the first-order discriminative signal vanishes \emph{iff} $\Delta(I) \perp g(\mathbf{b}(I))$. Caption-invariance delivers the factorisation, not the orthogonality, which rests on two further hypotheses: \textbf{(H1)} \emph{support disjointness}, $g(\mathbf{b}(I))$ lives on image-patch positions during prefill, disjoint from the caption positions where $\Delta(I)$ is non-negligible at leading order, so $\langle\Delta, g(\mathbf{b})\rangle$ is carried only by attention-mediated coupling; \textbf{(H2)} \emph{weak alignment across images}, $|\mathbb{E}_I\langle\Delta(I), g(\mathbf{b}(I))\rangle| \ll \mathrm{std}_I\langle\Delta(I), g(\mathbf{b}(I))\rangle/\sqrt{B}$ at the batch sizes used.

\begin{proposition}[First-order gradient vanishes under caption-invariance]
\label{prop:suppression}\sloppy
Under caption-invariance together with H1 and H2, the first-order component of $\mathbb{E}_I[\partial \mathcal{L}_{\mathrm{pair}}/\partial\gamma]$ is not distinguishable from zero at the batch sizes used, and the gate receives no usable discriminative signal from $\mathcal{L}_{\mathrm{pair}}$.
\end{proposition}

\noindent H1 and H2 are empirical claims about an architecture, so we measure them rather than assume them. On the RMS-prefix checkpoint, over $n{=}120$ SugarCrepe caption pairs with central differences at $\gamma{=}0$, the discriminative gradient has $|t|{=}0.19$ (mean $+0.674$, s.e.\ $3.512$), while the \emph{same} probe on the \emph{same} checkpoint resolves the single-caption cross-entropy gradient at $|t|{=}2.42$ (mean $-15.894$, s.e.\ $6.571$; full probe in the supplementary material). The null belongs to the paired objective, not to the measurement, which is the asymmetry caption-invariance predicts.

\noindent\textbf{Regime A: first-order dead $+$ second-order noise $\Rightarrow$ stall (S5).} When higher-order effects of the perturbation average to noise across images, the gradient is unbiased and the optimiser drifts without direction; $\gamma$ hovers near zero for the whole run and ends at $+0.005$.

\noindent\textbf{Regime B: first-order dead $+$ second-order systematic $\Rightarrow$ dive (S3).} When the perturbation systematically hurts the loss, the second-order gradient acquires a consistent sign that drives $\gamma$ toward its magnitude-zero limit, unbounded under softplus, hence the monotonic dive to $-8.5$. Two facts characterise this regime without a general theorem. For un-normalised injected tokens entering a pre-norm block, the first RMSNorm scales $\partial\,\mathrm{RMSNorm}(\mathbf{e}+\gamma\mathbf{t})/\partial\gamma$ at the origin by $\mathrm{rsqrt}(\varepsilon)$ and saturates once $\gamma\|\mathbf{t}\|$ exceeds $\sqrt{\varepsilon}$; with the measured token norms this is a ${\approx}316\times$ amplification collapsing beyond $|\gamma|\approx0.003$, a nonlinear well whose walls the optimiser meets long before any useful operating point. And S3 excludes capacity: with LoRA frozen, a monotone 50K-step drift is neither noise nor an adapter artefact. The curvature $\partial^2\mathcal{L}_{\mathrm{task}}/\partial\gamma^2|_0$ is not resolvable at $n{=}120$ ($|t|{=}0.02$); the regime is characterised, not proved.

\noindent\textbf{S1, S2, S4 sit in an intermediate regime.} Under the task cross-entropy the first-order gradient through $\gamma$ need not vanish, and suppression reflects a mild second-order preference for small $|\gamma|$; consistent with this, they settle at modest negatives ($-0.70$, $-0.60$, $-0.38$) rather than diving.

\begin{table}[t]
\centering
\caption{$2{\times}2$ factorisation, norm treatment $\times$ gate, with the pathway \emph{on} / \emph{zeroed} in every cell (grid proposals; Sec.~\ref{sec:protocol}). ``None'' = ungated prefix at scale 1 vs.\ 0 (HyperVis checkpoint~\cite{hypervis2026} / the ungated RMS prefix); ``Tanh'' = learned gate vs.\ 0 (S5 / RMS-gated base run). $\Delta$ = on $-$ off with paired 95\% CI. $^{a}$First 2{,}000 GQA testdev questions and SugarCrepe pairs. $^{b}$First 1{,}000 relational GQA questions and SugarCrepe pairs.}
\label{tab:2x2}
\vspace{-0.8em}
\resizebox{\columnwidth}{!}{
\begin{tabular}{@{}ll cl cl@{}}
\toprule
Norm & Gate & GQA on\,/\,off & $\Delta$ [95\% CI] & SC on\,/\,off & $\Delta$ [95\% CI] \\
\midrule
Raw & None (forced, $s{=}1$ vs $0$)$^{a}$ & 50.4 / 59.0 & $-8.6\ [-10.5, -6.6]$ & 79.9 / 73.4 & $+6.5\ [+5.1, +8.0]$ \\
Raw & Tanh (learned $\gamma$ vs $0$)$^{b}$ & 44.0 / 50.7 & $-6.7\ [-8.4, -5.0]$ & 80.0 / 78.5 & $+1.5\ [+0.7, +2.4]$ \\
RMS & None (forced, $s{=}1$ vs $0$)$^{a}$ & 48.5 / 60.8 & $-12.3\ [-14.1, -10.5]$ & 85.2 / 75.8 & $+9.3\ [+7.8, +10.8]$ \\
RMS & Tanh (learned $\gamma$ vs $0$)$^{b}$ & 50.2 / 49.9 & $+0.3\ [-0.5, +1.2]$ & 80.7 / 79.4 & $+1.3\ [+0.6, +2.1]$ \\
\bottomrule
\end{tabular}
}
\end{table}

\subsection{Behavioural Ablation: Closed Is Not Always Unused}
\label{sec:knockout}

\begin{table}[t]
\centering
\caption{Behavioural ablation under one protocol (grid proposals, single implementation). \emph{Off}: bias ${\times}0$ (S1--S3) or prefix zeroed with slots kept (S4--S5); \emph{on}: learned bias scale or gate; \emph{absent}: no prefix slots. $\Delta$ = on $-$ off, paired 95\% CI; flips = share of predictions that change. GQA-rel and SugarCrepe: first 1{,}000 items; GQA testdev: full set (S1--S2) or first 2{,}000 questions (S4--S5).}
\label{tab:ablations_main}
\vspace{-0.8em}
\resizebox{\columnwidth}{!}{
\begin{tabular}{@{}llrrrrrr@{}}
\toprule
 & & & \multicolumn{2}{c}{pathway} & & & pathway \\
\cmidrule(lr){4-5}
Design & learned gate & benchmark ($n$) & off & on & $\Delta$ on$-$off [95\% CI] & flips \% & absent \\
\midrule
S1 attn.\ bias ($\alpha_{\max}{=}0.05$) & $\beta{=}-0.696$ & GQA-rel (1,000) & 51.7 & 52.2 & $+0.5\ [-0.4, +1.3]$ & 4.4 & -- \\
 &  & SugarCrepe (1,000) & 77.4 & 77.2 & $-0.2\ [-0.7, +0.3]$ & 0.6 & -- \\
 &  & GQA testdev (12,578) & 57.93 & 58.55 & $+0.62\ [+0.39, +0.86]$ & 4.0 & -- \\
\midrule
S2 attn.\ bias ($\alpha_{\max}{=}0.01$) & $\beta{=}-0.600$ & GQA-rel (1,000) & 53.2 & 53.2 & $+0.0\ [-0.6, +0.5]$ & 2.3 & -- \\
 &  & SugarCrepe (1,000) & 78.8 & 78.7 & $-0.1\ [-0.6, +0.3]$ & 0.5 & -- \\
 &  & GQA testdev (12,578) & 60.09 & 60.14 & $+0.06\ [-0.15, +0.26]$ & 2.6 & -- \\
\midrule
S3 margin loss, LoRA frozen & $\beta{=}-8.496$ & GQA-rel (1,000) & 53.1 & 53.3 & $+0.2\ [-0.3, +0.8]$ & 1.5 & -- \\
 &  & SugarCrepe (1,000) & 77.2 & 77.3 & $+0.1\ [-0.3, +0.5]$ & 0.5 & -- \\
\midrule
S4 gated prefix, no dropout & $\tanh\gamma{=}-0.362$ & GQA-rel (1,000) & 50.5 & 50.6 & $+0.1\ [-1.1, +1.3]$ & 9.7 & 53.4 \\
 &  & SugarCrepe (1,000) & 80.1 & 81.5 & $+1.4\ [+0.5, +2.3]$ & 2.2 & 77.1 \\
 &  & GQA testdev (2,000) & 58.6 & 59.6 & $+0.9\ [+0.1, +1.8]$ & 8.3 & 60.1 \\
\midrule
S5 raw prefix, gated & $\tanh\gamma{=}+0.005$ & GQA-rel (1,000) & 50.7 & 44.0 & $-6.7\ [-8.4, -5.0]$ & 19.5 & 54.3 \\
 &  & SugarCrepe (1,000) & 78.5 & 80.0 & $+1.5\ [+0.7, +2.4]$ & 1.9 & 76.8 \\
 &  & GQA testdev (2,000) & 59.4 & 54.1 & $-5.2\ [-6.3, -4.2]$ & 13.6 & 60.6 \\
\bottomrule
\end{tabular}
}
\end{table}

Table~\ref{tab:ablations_main} ablates each pathway with real proposals. For S1, S2 and S3 the bias moves relational GQA by at most $+0.5$pp and SugarCrepe by at most $0.2$pp with every interval covering zero, and multiplying the bias by 3 changes nothing either (Table~\ref{tab:backbones}); on the full 12{,}578-question testdev set the S1 bias is worth $+0.62$pp $[0.39, 0.86]$, resolvable at that $n$ but small, and S2 is unchanged ($+0.06$pp [-0.15, 0.26]). The bias changes 2--4\% of predictions, balanced in direction, against 10--29\% disagreement between genuinely different models. The S4 prefix and every other RMS-normalised gated prefix move accuracy by at most 1.5pp at the learned gate (per-checkpoint ablations in the supplementary material); a few of these shifts are resolvable (S4 gains $1.4$pp $[0.5, 2.3]$ on SugarCrepe), so ``behaviourally closed'' means small, not exactly zero. Zeroed slots are not neutral either: for the RMS designs trained with LoRA on q/v they cost 2--5pp of relational GQA and gain 2--3pp of SugarCrepe relative to absent slots (the two LoRA-all-linear runs show no slot cost), and forcing the content fully on returns both to the absent level, so the learned gate parks these models at the zero-slot point.

\noindent\textbf{S5 is the exception, and it changes the reading of that design.} At its learned gate of $+0.005$ the un-normalised prefix costs $6.7$pp $[5.0, 8.4]$ on relational GQA relative to the zeroed slot and $5.2$pp on the 2{,}000-question testdev subset; the effect is sign-dependent ($\gamma{=}{-}0.005$ \emph{improves} relational GQA by $2.8$pp $[1.5, 4.2]$), grows with $|\gamma|$ (36.4\% at $\gamma{=}0.01$, 12.2\% at $\gamma{=}1$), and helps log-likelihood scoring ($+1.5$pp $[0.7, 2.4]$ on SugarCrepe at the learned gate, $+6.8$pp at $\gamma{=}0.03$). A gate that stalls at $+0.005$ can therefore still admit a signal the model uses, harmfully for generation and helpfully for scoring, because the RMSNorm well of Sec.~\ref{sec:regimes} makes the first $10^{-3}$ of gate range count for most of the effect. ``The gate always closes'' is a statement about the parameter; whether the pathway is used has to be measured.

\noindent\textbf{Attention mass cannot indicate use either.} Across all 32 decoder layers, the injected slots of the RMS-gated prefix draw ${\sim}40\times$ chance attention at the learned gate, ${\sim}38\times$ when the slots hold exact zero vectors, and only ${\sim}2\times$ when the scale-matched content is forced fully on; the soft prompt and the raw prefix behave alike at zero and at the learned gate (layer-wise numbers in the supplementary material). The mass is a property of the slot, a sink, not of the signal; a pathway can be attended at every depth and still unused, so we rely on ablations alone.

\section{Norm Mismatch and the Engineering Fix}
\label{sec:norm_fix}

When prefix tokens are derived from a hyperbolic manifold and projected into LLM embedding space, the resulting embeddings can carry L2 norms very far from those of the pretrained text embeddings. In the unnormalised configuration of~\cite{hypervis2026}, the raw prefix embeddings sit at $L2{\sim}24.6$ while text embeddings sit at $L2{\sim}0.68$, a \textbf{36$\times$ gap}, and generative VQA accuracy collapses (Table~\ref{tab:2x2}). We attribute the collapse to \emph{residual-stream dominance}, not to attention-softmax saturation.

LLaMA's pre-norm architecture applies RMSNorm before every Q/K/V projection, so the 36$\times$ input-embedding gap does not become a 36$\times$ key-norm gap at the softmax, and attention cannot saturate through this channel. What survives is the residual pathway: each layer adds a bounded block output to the running residual, so a prefix position that starts 36$\times$ larger stays prefix-dominated at every subsequent layer, and the unembedding reads out a distribution skewed by the prefix subspace wherever prefix tokens are attended.

\noindent\textbf{Why a raw prefix helps compositional scoring while hurting generation.} Log-likelihood scoring compares two captions conditioned on the \emph{same} image-prefix pair, so the residual-dominance bias largely cancels and whatever compositional signal the prefix direction carries survives; autoregressive generation needs accurate next-token logits in absolute terms, which residual-stream skew disrupts. Table~\ref{tab:2x2} shows the trade in every un-normalised cell: switching the HyperVis prefix on with real proposals costs $8.6$pp $[6.6, 10.5]$ of GQA on the 2{,}000-question subset, while the S5 prefix at $\gamma{=}0.03$ gains $6.8$pp of SugarCrepe.

\begin{figure*}[t]
\centering
\includegraphics[width=.8\linewidth]{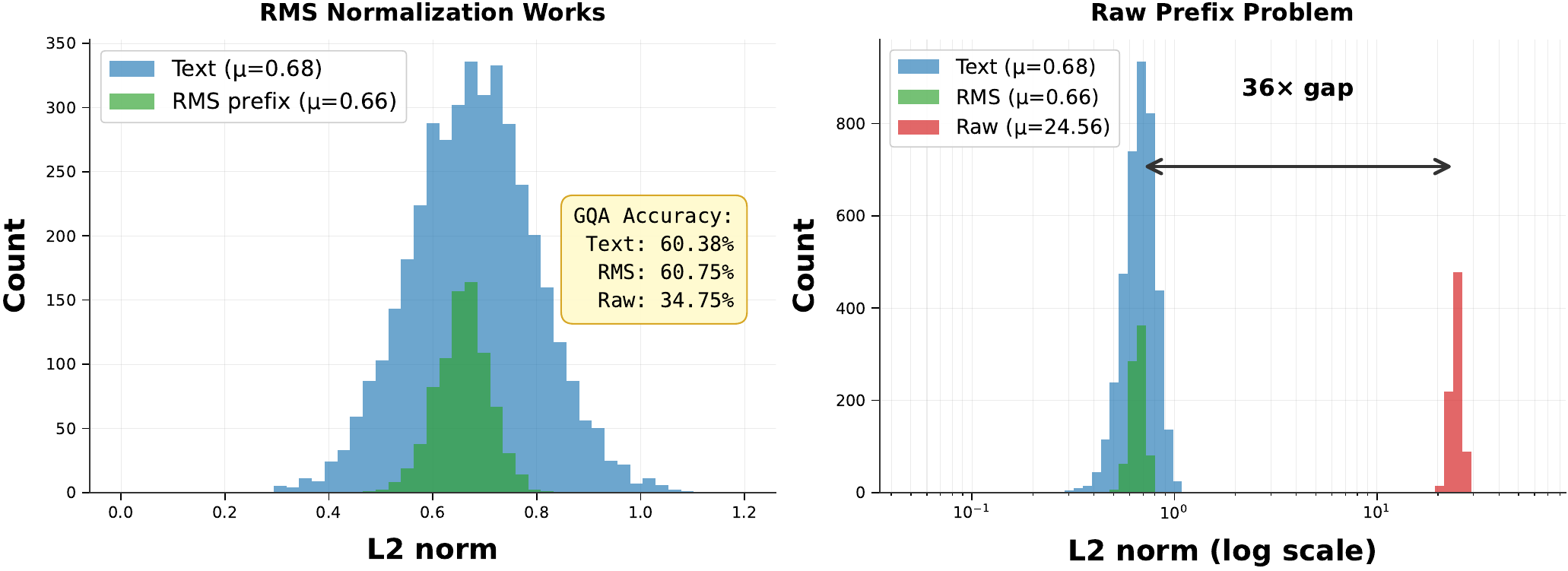}
\caption{L2 norm distributions for text embeddings vs.\ raw and RMS-normalised prefix tokens. Raw prefixes carry 36$\times$ the text-embedding norm, which dominates the residual stream and breaks autoregressive generation; RMS norm aligns the two distributions.}
\label{fig:norms}
    
\end{figure*}

\noindent\textbf{The fix: RMS normalization.} We apply the RMSNorm-and-text-mean rescaling of Eq.~\eqref{eq:prefix} post-projection to every prefix token so that the resulting L2-norm distribution tracks the text embeddings (Fig.~\ref{fig:norms}).

\noindent\textbf{$2{\times}2$ factorisation.} Table~\ref{tab:2x2} crosses norm treatment with gating, every cell measured on / zeroed. Norm alignment turns the un-gated prefix's collapse (HyperVis: $59.0\to50.4$) into a milder but still substantial cost ($60.8\to48.5$ for the RMS prefix at scale~1) while SugarCrepe gains $9.3$pp $[7.8, 10.8]$; at scale~0.3 the RMS prefix costs nothing on GQA ($61.5$ vs.\ $60.8$) and gains $1.7$pp on SugarCrepe (full scale sweep in the supplementary material). A learnable gate on either norm treatment closes in the parameter (raw: $\gamma\to+0.005$; RMS: $\gamma\to-0.15$), but only the RMS gate is also behaviourally closed; the raw one is the open-and-harmful S5. RMS alignment is therefore \emph{generation-preserving relative to a raw prefix}, not generation-safe: an RMS prefix that is actually on still costs GQA, and the recipe that keeps generation at the vanilla level is the one in which the prefix is dropped at inference (Sec.~\ref{sec:matched}).

\section{Experimental Results}
\label{sec:results}

\begin{table}[t]
\centering
\caption{Main results on full GQA testdev (12{,}578 questions) and SugarCrepe (7{,}514 pairs). GQA is measured with the auxiliary pathway absent at inference (the LoRA-only model each recipe produces); SugarCrepe with the pathway on and grid proposals. Gate: end-of-training value ($\beta$ for the attention bias, $\tanh\gamma$ for prefixes). Raw prefix row: values reported in~\cite{hypervis2026}. Winoground group scores sit at chance (16.75--19.75\% vs.\ 16.67\%) for all configurations.}
\label{tab:main}
\vspace{-0.8em}
\resizebox{\columnwidth}{!}{
\begin{tabular}{llccl}
\toprule
\textbf{Method} & \textbf{Gate} & \textbf{GQA} & \textbf{SC avg} & \textbf{Key finding} \\
\midrule
LLaVA-1.5-7B vanilla & --- & 60.38 & 73.69 & Baseline \\
\quad + LoRA only (no geo.\ loss) & --- & 57.21 & 74.98 & LoRA alone degrades GQA \\
\quad + Raw prefix~\cite{hypervis2026} & --- & 61.03 & 79.94 & Prefix on: 34.75 reported, 50.4 w/ proposals \\
\midrule
\quad + Attn.\ bias (small $\alpha$) & $-0.60$ & 60.16 & 75.73 & Bias inert at inference \\
\quad + RMS prefix, ungated & forced & \textbf{60.75} & \textbf{84.90} & Prefix on: GQA 48.5 (Tab.~\ref{tab:2x2}) \\
\quad + Soft prompt (no geometry) & $-0.20$ & 59.90 & 77.00 & Content-free; strong SC \\
\quad + Euclidean gated (no geo.\ loss) & $+0.19$ & 59.73 & 76.56 & Gate opens, pathway inert \\
\quad + Raw prefix, gated (S5) & $+0.005$ & 60.00 & 76.18 & Open and harmful (Tab.~\ref{tab:ablations_main}) \\
\quad + Hyp.\ prefix, \emph{no geo.\ loss} (G2) & $-0.31$ & 59.59 & 78.84 & Best gated SC; no geometry \\
\quad + Soft prompt, 2nd run & $+0.12$ & 60.07 & 77.45 & Soft-prompt replicate \\
\bottomrule
\end{tabular}}
\end{table}

Table~\ref{tab:main} lists the headline numbers: GQA with each recipe's pathway absent, SugarCrepe with it on. Three readings follow. LoRA fine-tuning on GQA without any auxiliary signal costs 3pp of GQA, whereas every recipe trained with an auxiliary pathway, geometric or not, lands within 0.8pp of vanilla once that pathway is dropped. On SugarCrepe the ungated RMS prefix, on, scores highest (84.90\%) while the same setting costs 12pp of GQA (Table~\ref{tab:2x2}); among the gated recipes the two highest averages come from configurations with no geometry (G2 78.84\%, a second soft-prompt run 77.45\%), above the first soft prompt and every geometric variant. Per category (supplementary material), the RMS prefix on tops every column and, among the gated recipes, G2 wins all seven categories, so we do not read SugarCrepe as evidence of relational structure.

\subsection{Matched Inference Mode: Injecting Hurts, Dropping Recovers}
\label{sec:matched}

\begin{table*}[t]
\centering
\caption{Every configuration under one inference mode. \emph{Drop}: pathway removed at inference (LoRA-only); \emph{on}: pathway active (prefix at learned gate, or scale 1 for the ungated RMS prefix; grid proposals). $\Delta$ = drop $-$ on with paired 95\% CI (10{,}000 resamples on shared items). VSR~\cite{liu2023vsr} is out of distribution. \emph{r1}/\emph{r2}: the RMS-prefix checkpoint of Table~\ref{tab:main} and a second run of the same recipe; \emph{geo.}: trained with the geometric losses.}
\label{tab:matched}
\vspace{-0.8em}
\resizebox{\textwidth}{!}{
\begin{tabular}{@{}l rrl rrl rrl@{}}
\toprule
 & \multicolumn{3}{c}{GQA-rel ($n{=}5{,}308$)} & \multicolumn{3}{c}{GQA-attr ($n{=}5{,}186$)} & \multicolumn{3}{c}{VSR ($n{=}2{,}113$)} \\
\cmidrule(lr){2-4}\cmidrule(lr){5-7}\cmidrule(lr){8-10}
Config & drop & on & $\Delta$ [95\% CI] & drop & on & $\Delta$ [95\% CI] & drop & on & $\Delta$ [95\% CI] \\
\midrule
LLaVA-1.5 vanilla & 52.9 & -- & -- & 66.1 & -- & -- & 66.2 & -- & -- \\
RMS prefix r1 (hyp., geo.) & 53.4 & 37.9 & $+15.5\ [14.3, 16.7]$ & 66.6 & 59.3 & $+7.4\ [6.2, 8.6]$ & 66.5 & 55.6 & $+10.9\ [9.0, 12.9]$ \\
RMS prefix r2 (hyp., geo.) & 53.3 & 45.9 & $+7.4\ [6.3, 8.4]$ & 66.2 & 62.4 & $+3.9\ [2.8, 5.0]$ & 68.5 & 56.2 & $+12.3\ [10.0, 14.6]$ \\
Euclid.\ containment (flat, geo.) & 53.2 & 52.4 & $+0.8\ [0.2, 1.5]$ & 66.7 & 68.0 & $-1.2\ [-2.2, -0.3]$ & 63.8 & 56.6 & $+7.2\ [5.8, 8.7]$ \\
soft prompt & 53.0 & 50.1 & $+2.9\ [2.1, 3.8]$ & 65.1 & 67.2 & $-2.1\ [-3.2, -1.0]$ & 63.6 & 56.0 & $+7.6\ [6.1, 9.1]$ \\
Euclid.\ gated (no geo.) & 52.3 & 49.6 & $+2.7\ [1.9, 3.5]$ & 65.3 & 67.0 & $-1.7\ [-2.7, -0.7]$ & 62.8 & 56.0 & $+6.9\ [5.4, 8.3]$ \\
G2 hyp.\ prefix (no geo.) & 52.3 & 49.5 & $+2.8\ [1.9, 3.7]$ & 65.1 & 67.1 & $-2.0\ [-3.1, -0.9]$ & 63.4 & 55.4 & $+8.0\ [6.5, 9.5]$ \\
attn.\ bias (small $\alpha$) & 52.4 & 52.6 & $-0.1\ [-0.4, 0.2]$ & 67.5 & 67.5 & $+0.0\ [-0.3, 0.3]$ & 54.5 & 54.5 & $-0.0\ [-0.2, 0.1]$ \\
\bottomrule
\end{tabular}
}
\end{table*}

Comparing geometric configurations with the pathway \emph{dropped} against no-geometry configurations with it \emph{active} would credit the geometry with about 3pp of relational GQA that it does not have; under matched modes (Table~\ref{tab:matched}) three results hold.

\noindent\textbf{Dropped, no configuration separates from vanilla in-distribution.} With the pathway removed, no configuration separates from vanilla on relational GQA (paired $\Delta$ from $-0.62$ to $+0.47$pp, all intervals covering zero; paired contrasts in the supplementary material); on attribute questions only the attention-bias recipe does ($+1.35$pp $[0.46, 2.26]$). Out of distribution the dropped models split: the two hyperbolic RMS prefixes sit at or above vanilla on VSR ($+0.3$ and $+2.3$pp), the flat, soft-prompt and G2 recipes 2.4--3.4pp below it, and the attention-bias recipe 11.7pp below. Training-time geometric losses, hyperbolic or flat, give no measurable in-distribution gain once the pathway is dropped, so they do not act as a regulariser of LoRA in the sense a dropped-pathway recipe would need.

\noindent\textbf{Active, every prefix pathway costs the reasoning it targets.} Keeping the trained pathway on costs $2.7$--$2.9$pp of relational GQA for the content-free and no-loss prefixes (soft prompt, Euclidean, G2), $0.8$pp for the flat containment prefix, and $7.4$ and $15.5$pp for the two RMS geometric prefixes, all with intervals excluding zero: the harder a checkpoint injects, the more injection hurts. Out of distribution the picture is the same and larger: with the pathway on, every prefix configuration collapses to 55.4--56.6\% VSR, and dropping it recovers $6.9$--$12.3$pp. The attention-bias pathway is the one design for which on and dropped coincide.

\noindent\textbf{Injection trades axes.} The same active prefixes that hurt relational questions \emph{help} attribute questions by $1.2$--$2.1$pp (soft prompt, Euclidean, G2, flat containment), while the RMS geometric prefixes are hurt on both axes. Where an injected caption-invariant pathway changes behaviour at all, it shifts capacity from relational and spatial questions toward attribute questions. On the axes a relational signal is meant to serve, dropping it wins.

\subsection{Geometry: The Objective, Not the Manifold}
\label{sec:geometry}

\textbf{Is the hyperbolic manifold necessary? Not in-distribution.} The Euclidean containment control (flat-space graph, same $\mathcal{L}_{\text{ent}}$ and $\mathcal{L}_{\text{ang}}$ on cosine angles) lands at 53.18\% relational GQA with the pathway dropped, against 53.41\% and 53.28\% for the two hyperbolic checkpoints (paired $\Delta{=}{+}0.23$pp $[-0.38, +0.83]$, $p{=}0.49$, and $+0.09$pp $[-0.60, +0.77]$, $p{=}0.84$); across one-epoch seeds (two hyperbolic, three flat) the gap is $-0.10$pp $[-0.55,+0.35]$. Flat space ends training with a $61\times$ higher containment loss ($0.662$ vs.\ $0.0108$) and the same accuracy, so it is not the satisfied constraint that matters. Whether the containment \emph{objective} matters is a small effect: under matched dropped-pathway evaluation the flat graph with containment losses beats the flat graph without them by $+0.87$pp $[0.34, 1.41]$ on relational GQA and $+1.37$pp $[0.69, 2.04]$ on attribute questions; a mixed-mode comparison would put it at $+3.5$pp. Out of distribution there is a hyperbolic \emph{trend}: three hyperbolic seeds reach $67.9\pm1.2\%$ VSR with the pathway dropped against $65.4\pm1.5\%$ for three flat seeds and 66.2\% for vanilla (Welch $p{=}0.09$ and $0.14$; per-seed numbers in the supplementary material). Paired contrasts between individual checkpoints do separate ($+2.7$ and $+4.7$pp), but checkpoints differ by training seed, so the seed-level test is the right one; the trend is consistent but underpowered, and we do not claim it.

\subsection{Generation Safety: Hallucination and Fluency}
\label{sec:generation_safety}

We evaluate POPE (yes/no object existence on COCO, $3{\times}3{,}000$ questions) and COCO val2017 captioning (5{,}000 images) for vanilla, the attention-bias model with its bias on, the RMS-prefix model and the soft prompt; the full table is in the supplementary material. The RMS-prefix entries evaluate that recipe with the prefix zeroed (slots kept), the reference condition of Sec.~\ref{sec:protocol}. Under it, the recipe costs 1--2pp of POPE relative to vanilla (adversarial 82.30 vs.\ 83.40) and 3\% of CIDEr (1.096 vs.\ 1.132), soft-prompt tuning gives up 13\% of CIDEr (0.987), and the attention-bias model, bias on, matches vanilla on POPE (84.13 adversarial) at a 6\% CIDEr cost. Training the RMS recipe with 50\% prefix dropout restores POPE to the vanilla level ($86.40/85.70/84.07$, pathway absent). With the RMS prefix genuinely on at scale~1, POPE is $85.5/84.8/83.4$, within 0.8pp of vanilla and 1.1--1.5pp \emph{above} the zeroed condition, whereas the same checkpoint loses 12.3pp of GQA (Table~\ref{tab:2x2}): injection is harmless on existence questions and harmful on relational ones, and the recipe that keeps every benchmark at vanilla level drops the prefix.

\section{Conclusion}
\label{sec:conclusion}

We treated learnable gates on auxiliary VLM pathways as objects of study. Under cross-entropy the gate closes in every design and backbone we trained, for a measured first-order dead-gradient reason and a characterised second-order negative-utility reason. Closure of the parameter says nothing about use: ablated with real proposals, the norm-aligned pathways are inert, the un-normalised one is used and harmful, and attention onto the injected slots is a sink either way. Under one inference mode, injecting the trained pathway costs relational and out-of-distribution spatial accuracy for every prefix design, training-time geometry alone gives no in-distribution gain, and flat containment ties the hyperbolic manifold; norm alignment keeps a prefix from collapsing generation, and dropping it keeps generation at the vanilla level.

\noindent\textbf{Scope and limitations.} The claims are scoped to caption-invariant, image-derived pathways gated by a learnable scalar under a task-dominant objective, on three 7--8B instruction-tuned VLMs of which only LLaVA-1.5 carries all five designs. Most LLaVA configurations are single training runs; the training-seed standard deviation on five matched runs is ${\approx}0.2$pp on relational GQA against effects of 2.7--15.5pp. Text-coupled pathways, which break the caption-invariance that nulls the first-order gradient, are the natural next step. The recommendation is narrow and testable: for a caption-invariant image-derived signal, a learnable gate reports nothing about use, so test use by ablation with real inputs; if the signal must be injected, inject it at matched norm without an optional gate and expect a relational cost unless the objective rewards the pathway; otherwise drop the pathway at inference.

\bibliographystyle{splncs04}
\bibliography{main}

@String(CVPR  = {CVPR})

@String(ICCV  = {ICCV})

@String(ECCV  = {ECCV})

@String(NeurIPS = {NeurIPS})

@String(ICML  = {ICML})

@String(ICLR  = {ICLR})

@String(ICASSP=	{ICASSP})

@String(EMNLP = {EMNLP})

@inproceedings{liu2024llava15,
  title={Improved Baselines with Visual Instruction Tuning},
  author={Liu, Haotian and Li, Chunyuan and Li, Yuheng and Lee, Young Jae},
  booktitle=CVPR,
  year={2024}
}

@article{wang2024qwen2vl,
  title={{Qwen2-VL}: Enhancing Vision-Language Model's Perception of the World at Any Resolution},
  author={Wang, Peng and others},
  journal={arXiv:2409.12191},
  year={2024}
}

@inproceedings{yuksekgonul2023aro,
  title={When and Why Vision-Language Models Behave like Bags-of-Words, and What to Do About It?},
  author={Y{\"u}ksekgonul, Mert and Bianchi, Federico and Kalluri, Pratyusha and Jurafsky, Dan and Zou, James},
  booktitle=ICLR,
  year={2023}
}

@inproceedings{thrush2022winoground,
  title={Winoground: Probing Vision and Language Models for Visio-Linguistic Compositionality},
  author={Thrush, Tristan and Jiang, Ryan and Bartolo, Max and others},
  booktitle=CVPR,
  year={2022}
}

@inproceedings{hsieh2024sugarcrepe,
  title={{SugarCrepe}: Fixing Hackable Benchmarks for Vision-Language Compositionality},
  author={Hsieh, Cheng-Yu and Zhang, Jieyu and Ma, Zixian and others},
  booktitle=NeurIPS,
  year={2023}
}

@inproceedings{nickel2017poincare,
  title={Poincar{\'e} Embeddings for Learning Hierarchical Representations},
  author={Nickel, Maximilian and Kiela, Douwe},
  booktitle=NeurIPS,
  year={2017}
}

@inproceedings{ganea2018hyperbolicnn,
  title={Hyperbolic Entailment Cones for Learning Hierarchical Embeddings},
  author={Ganea, Octavian-Eugen and B{\'e}cigneul, Gary and Hofmann, Thomas},
  booktitle=ICML,
  year={2018}
}

@inproceedings{desai2023meru,
  title={Hyperbolic Image-Text Representations},
  author={Desai, Karan and Nickel, Maximilian and Rajpurohit, Tanmay and Johnson, Justin and Vedantam, Ramakrishna},
  booktitle=ICML,
  year={2023}
}

@inproceedings{pal2025hycoclip,
  title={Compositional Entailment Learning for Hyperbolic Vision-Language Models},
  author={Pal, Avik and others},
  booktitle=ICLR,
  year={2025}
}

@inproceedings{ramasinghe2024accept,
  title={Accept the Modality Gap: An Exploration in the Hyperbolic Space},
  author={Ramasinghe, Sameera and Khan, Salman and Barnes, Nick and others},
  booktitle=CVPR,
  year={2024}
}

@book{ungar2005analytic,
  title={Analytic Hyperbolic Geometry: Mathematical Foundations and Applications},
  author={Ungar, Abraham A},
  year={2005},
  publisher={World Scientific}
}

@inproceedings{press2022alibi,
  title={Train Short, Test Long: Attention with Linear Biases Enables Input Length Generalization},
  author={Press, Ofir and Smith, Noah A and Lewis, Mike},
  booktitle=ICLR,
  year={2022}
}

@article{su2024rope,
  title={{RoFormer}: Enhanced Transformer with Rotary Position Embedding},
  author={Su, Jianlin and Lu, Yu and Pan, Shengfeng and Murtadha, Ahmed and Liu, Bo and Liu, Yunfeng},
  journal={Neurocomputing},
  year={2024}
}

@inproceedings{wang2024llavassg,
  title={Enhancing Visual Grounding and Generalization: A Multi-Task Learning Approach with Scene Graph Expressions},
  author={Wang, Henghui and others},
  booktitle=ICASSP,
  year={2025}
}

@inproceedings{mitra2024ccot,
  title={Compositional Chain-of-Thought Prompting for Large Multimodal Models},
  author={Mitra, Chancharik and Huang, Brandon and Darrell, Trevor and Herzig, Roei},
  booktitle=CVPR,
  year={2024}
}

@inproceedings{herzig2023structured,
  title={Incorporating Structured Representations into Pretrained Vision \& Language Models Using Scene Graphs},
  author={Herzig, Roei and others},
  booktitle=EMNLP,
  year={2023}
}

@inproceedings{jia2022vpt,
  title={Visual Prompt Tuning},
  author={Jia, Menglin and Tang, Luming and Chen, Bor-Chun and others},
  booktitle=ECCV,
  year={2022}
}

@article{gao2023llamaadapter,
  title={{LLaMA-Adapter V2}: Parameter-Efficient Visual Instruction Model},
  author={Gao, Peng and Han, Jiaming and Zhang, Renrui and others},
  journal={arXiv:2304.15010},
  year={2023}
}

@inproceedings{peng2025hyperet,
  title={Efficient Training in Hyperbolic Space for Multi-modal Large Language Models},
  author={Peng, Lingyu and Xu, Minghao and Liu, Chang and Yang, Yi and Shen, Zhuowen},
  booktitle=NeurIPS,
  year={2025}
}

@article{he2025hyperbolic_survey,
  title={Hyperbolic Deep Learning for Foundation Models: A Survey},
  author={He, Menglin and others},
  journal={SIGKDD},
  year={2025}
}

@inproceedings{dalal2026attwarp,
  title={Constructive Distortion for Vision-Language Models via Attention-Guided Image Warping},
  author={Dalal, Harneet Singh and Kashiani, Hamed and Afghah, Fatemeh},
  booktitle=ICLR,
  year={2026}
}

@inproceedings{mishra2025scramble,
  title={{SCRAMBLe}: Preference-Tuned Vision-Language Models for Visio-Linguistic Compositionality},
  author={Mishra, Samarth and Saenko, Kate and Saligrama, Venkatesh},
  booktitle=ICCV,
  year={2025}
}

@inproceedings{yang2026scenegraph_thinking,
  title={Scene Graph Thinking: Reinforcing Structured Visual Reasoning for Multimodal Large Language Models},
  author={Yang, Chengzu and others},
  booktitle=ICML,
  year={2026}
}

@inproceedings{wu2025flashbias,
  title={{FlashBias}: Fast Computation of Attention with Bias},
  author={Wu, Yichao and others},
  booktitle=NeurIPS,
  year={2025}
}

@inproceedings{dumpala2024sugarcrepepp,
  title={{SugarCrepe++}: Fixing Compositionality Benchmarks with Hard Positives},
  author={Dumpala, Sri Harsha and others},
  booktitle={NeurIPS D\&B},
  year={2024}
}

@article{liu2023vsr,
  title={Visual Spatial Reasoning},
  author={Liu, Fangyu and Emerson, Guy and Collier, Nigel},
  journal={Transactions of the Association for Computational Linguistics},
  volume={11},
  pages={635--651},
  year={2023}
}

@inproceedings{jain2024neftune,
  title={{NEFTune}: Noisy Embeddings Improve Instruction Finetuning},
  author={Jain, Neel and Chiang, Ping-yeh and Wen, Yuxin and Kirchenbauer, John and Chu, Hong-Min and Somepalli, Gowthami and Bartoldson, Brian R and Kailkhura, Bhavya and Schwarzschild, Avi and Saha, Aniruddha and Goldblum, Micah and Geiping, Jonas and Goldstein, Tom},
  booktitle={ICLR},
  year={2024}
}

@article{hypervis2026,
  title={{HyperVis}: Continuous Latent Visual Relational Graphs on the {Lorentz} Hyperboloid for Compositional Reasoning},
  author={Farazi, Moshiur and Ramasinghe, Sameera and Turza, Mahbub Ahmed and Rahman, Shafin},
  journal={arXiv:2606.06100},
  year={2026}
}

@software{geoopt,
  title={Geoopt: Riemannian Optimization in {PyTorch}},
  author={Kochurov, Max and Karimov, Rasul and Kozlukov, Sergei},
  year={2020},
  url={https://github.com/geoopt/geoopt}
}

@article{zhu2025internvl3,
  title   = {Intern{VL}3: Exploring Advanced Training and Test-Time Recipes for Open-Source Multimodal Models},
  author  = {Zhu, Jinguo and Wang, Weiyun and Chen, Zhe and Liu, Zhaoyang and Ye, Shenglong and Gu, Lixin and Tian, Hao and Duan, Yuchen and Su, Weijie and Shao, Jie and others},
  journal = {arXiv preprint arXiv:2504.10479},
  year    = {2025}
}

\clearpage
\setcounter{figure}{0}\setcounter{table}{0}\setcounter{section}{0}
\renewcommand{\thefigure}{S\arabic{figure}}\renewcommand{\theHfigure}{S\arabic{figure}}
\renewcommand{\thetable}{S\arabic{table}}\renewcommand{\theHtable}{S\arabic{table}}
\renewcommand{\thesection}{S\arabic{section}}\renewcommand{\theHsection}{S\arabic{section}}
\begin{center}{\Large\bfseries Supplementary Material}\end{center}
\medskip

\noindent This supplementary material accompanies the main paper above. Section, figure, and table numbers are prefixed with \textbf{S}; citations use the same bibliography as the main paper.

\medskip

\noindent\textbf{Guide to the Supplementary.}
\begin{center}
\scalebox{0.92}{
\begin{tabular}{@{}p{0.46\linewidth}p{0.46\linewidth}@{}}
\toprule
\textbf{Supplementary section} & \textbf{Extends this part of the main paper} \\
\midrule
\ref{supp:sec:related}: Extended Related Work & Sec.~1, positioning paragraph \\
\ref{supp:sec:geom}: Extended Geometric Primitives & Sec.~2.1--2.2 \\
\ref{supp:sec:training}: Training and Compute Details & Sec.~2.3 (Training setup) \\
\ref{supp:sec:ablations}: Behavioural Ablations of the Auxiliary Pathway (all designs, full sweeps) & Sec.~3.1, 3.5 and Table~3 \\
\ref{supp:sec:matched}: Matched Inference Mode: Contrasts, Seeds, Prefix Scale & Sec.~4, 5.1, 5.2 \\
\ref{supp:sec:probes}: Gate-Gradient and Attention Probes & Sec.~3.4, 3.5 \\
\ref{supp:sec:sc_breakdown}: Per-Category SugarCrepe Breakdown & Sec.~5, the SC average column \\
\ref{supp:sec:generation}: Generation-Side Evaluation (POPE, COCO) & Sec.~5.3 \\
\ref{supp:sec:discussion}: Extended Discussion & Sec.~6 \\
\ref{supp:sec:limitations}: Limitations & Sec.~6, scope paragraph \\
\bottomrule
\end{tabular}}
\end{center}

\medskip

\section{Extended Related Work}
\label{supp:sec:related}

The Introduction of the main paper lists the six threads closest to this work in one paragraph and points here for the fuller treatment. This section expands each thread and positions our contribution against it.

\noindent\textbf{Prefix and prompt tuning.}
VPT~\cite{jia2022vpt} introduced learned prefix tokens for vision transformers, and LLaMA-Adapter~\cite{gao2023llamaadapter} carried the idea into LLM layers with zero-initialised gating. In both cases the prefix content is learned end-to-end via backpropagation through the frozen backbone. Different to these works, we ask what happens when the prefix content is derived from a structured external computation --- here a geometric relational graph --- rather than shaped by the task gradient itself. We find that the gate on such a pathway closes under a task-only loss in every design we trained, and we measure separately whether the pathway is used.

\noindent\textbf{Auxiliary signal injection in VLMs.}
A growing body of work attaches scene-graph generators and feeds predicate triplets as text~\cite{wang2024llavassg,mitra2024ccot,herzig2023structured}, injects visual relation encodings as prefix tokens~\cite{hypervis2026}, or modulates attention weights with structural biases. Yang \etal~\cite{yang2026scenegraph_thinking} instead use scene graphs as reinforcement signals. These efforts differ in what they inject, but a common design assumption is that a well-shaped auxiliary signal will be consumed at inference. Our investigation spans two injection mechanisms (attention bias and prefix tokens) and characterises the conditions under which the pathways carrying that signal remain active or become suppressed.

\noindent\textbf{Hyperbolic VLMs.}
Hyperbolic spaces embed hierarchical structure with low distortion~\cite{nickel2017poincare,ganea2018hyperbolicnn,he2025hyperbolic_survey}. MERU~\cite{desai2023meru} introduced hyperbolic contrastive VLMs, and HyCoCLIP~\cite{pal2025hycoclip} extended this with compositional entailment. Ramasinghe \etal~\cite{ramasinghe2024accept} showed that geodesic contrastive losses conflict with entailment cones and cause curvature collapse; our angle-based losses inherit this insight. HyperET~\cite{peng2025hyperet} demonstrated efficient hyperbolic training for MLLMs. In our experiments the Lorentz manifold serves as the experimental vehicle for generating the auxiliary signal. The closure and norm-mismatch findings are mechanism-level rather than geometry-level; a Euclidean control that optimises the same containment objective matches the hyperbolic one in-distribution (main paper Sec.~5.2).

\noindent\textbf{Structured attention biases.}
ALiBi~\cite{press2022alibi} and RoPE~\cite{su2024rope} inject position-dependent biases into attention, and FlashBias~\cite{wu2025flashbias} provides efficient kernels for dense variants. Dalal \etal~\cite{dalal2026attwarp} improve LLaVA's compositionality through attention-guided image warping. Our attention-bias mechanism is a special case of this family; its gate closes and, with real proposals, the bias is behaviourally inert (main paper Table~3).

\noindent\textbf{Compositionality benchmarks.}
ARO~\cite{yuksekgonul2023aro}, Winoground~\cite{thrush2022winoground} and SugarCrepe~\cite{hsieh2024sugarcrepe} probe compositional understanding. SugarCrepe++ addresses known text-only biases in the original benchmark~\cite{dumpala2024sugarcrepepp}. SCRAMBLe~\cite{mishra2025scramble} reaches 54.8\% Winoground group via preference tuning. Our evaluation uses SugarCrepe (7 subcategories) and Winoground; we note the known biases where they bear on the interpretation.

\noindent\textbf{Auxiliary noise as regularisation.}
NEFTune~\cite{jain2024neftune} showed that adding uniform noise to input embeddings during instruction tuning improves downstream evaluation scores. This raises the possibility that part of the benefit attributed to structured auxiliary signals is captured by generic noise perturbation. The G2 control (same prefix architecture, geometric losses removed) addresses part of this confound; a pure random-tensor prefix remains future work (Sec.~\ref{supp:sec:limitations}).

\section{Extended Geometric Primitives}
\label{supp:sec:geom}

The main paper Sec.~2.1 uses several hyperbolic-geometry primitives (manifold, exponential map, entailment cone, central angle, Einstein midpoint) by name only. This section gives their explicit forms.

\noindent\textbf{Lorentz manifold.} The Lorentz model $\Ln$ is the upper sheet of the hyperboloid in $(n{+}1)$-dimensional Minkowski spacetime ($n{=}256$):
\begin{equation}
\Ln = \left\{\mathbf{p} \in \mathbb{R}^{n+1} : \linner{\mathbf{p}}{\mathbf{p}} = -\frac{1}{\kappa},\; p_0 > 0\right\},
\label{supp:eq:lorentz}
\end{equation}
where the Lorentzian inner product is $\linner{\mathbf{p}}{\mathbf{q}} = -p_0 q_0 + \sum_{i=1}^n p_i q_i$ and $\kappa > 0$ is the learnable curvature parameter (initialised at 1.0, clamped to $[0.1, 5.0]$). Tangent vectors at the origin have their time component fixed at $v_0=0$.

\noindent\textbf{Exponential map at the origin.} The exponential map at $\mathbf{o} = (\sqrt{1/\kappa}, 0, \ldots, 0)^\top$ sends a tangent vector $\mathbf{v}$ (with $v_0{=}0$) to a manifold point:
\begin{equation}
\exp_\mathbf{o}^\kappa(\mathbf{v}) = \cosh\!\left(\sqrt{\kappa}\|\tilde{\mathbf{v}}\|\right)\mathbf{o} + \frac{\sinh(\sqrt{\kappa}\|\tilde{\mathbf{v}}\|)}{\sqrt{\kappa}\|\tilde{\mathbf{v}}\|}\mathbf{v},
\label{supp:eq:expmap}
\end{equation}
followed by manifold projection $p_0 \leftarrow \sqrt{1/\kappa + \|\tilde{\mathbf{p}}\|^2}$, where $\tilde{\mathbf{p}} = (p_1, \ldots, p_n)$ denotes the spatial coordinates.

\noindent\textbf{Central angle.} Two points $\mathbf{p}, \mathbf{q}$ are separated by the central angle at the origin
\begin{equation}
\phi(\mathbf{p}, \mathbf{q}) = \arccos\!\left(\frac{\langle \tilde{\mathbf{u}}_\mathbf{p},\, \tilde{\mathbf{u}}_\mathbf{q} \rangle}{\|\tilde{\mathbf{u}}_\mathbf{p}\|\;\|\tilde{\mathbf{u}}_\mathbf{q}\|}\right),
\label{supp:eq:angle}
\end{equation}
where $\tilde{\mathbf{u}}_\mathbf{p}$ is the spatial part of $\log_\mathbf{o}^\kappa(\mathbf{p})$.

\noindent\textbf{Entailment cone half-aperture.} Every point $\mathbf{q}\in\Ln$ with $\|\tilde{\mathbf{q}}\|\ge K$ defines an entailment cone with half-aperture
\begin{equation}
\omega(\mathbf{q}) = \arcsin\!\left(\frac{K}{\|\tilde{\mathbf{q}}\|}\right), \qquad K{=}0.1.
\label{supp:eq:aperture}
\end{equation}
The cone containment test $\phi(\mathbf{p},\mathbf{q}) \le \omega(\mathbf{q})$ is inherently asymmetric because different points carry different apertures --- points closer to the origin have larger $\omega(\mathbf{q})$ and therefore contain more of the manifold, matching the intended semantics that general concepts contain specific ones.

\noindent\textbf{Einstein midpoint (aggregation).} Per-node embeddings are computed from per-edge relation tangents via the Einstein midpoint~\cite{ungar2005analytic}. Denoting the manifold-mapped edge embeddings for node $i$ as $\{\mathbf{z}_{ij}\}_{j=1}^N$ with time components $\gamma_{ij} = z_{ij,0}$, the weighted Minkowski sum is
\begin{equation}
\mathbf{z}_i^{\text{raw}} = \frac{\sum_{j=1}^N \gamma_{ij}\,\mathbf{z}_{ij}}{\left|\sum_{j=1}^N \gamma_{ij}\right|},
\qquad \mathbf{z}_i = \text{proj}_{\Ln}\!\big(\mathbf{z}_i^{\text{raw}}\big),
\label{supp:eq:einstein}
\end{equation}
where the projection restores $z_{i,0} = \sqrt{1/\kappa + \|\tilde{\mathbf{z}}_i^{\text{raw}}\|^2}$. Weighting by the time component (rather than uniformly) is what distinguishes the Einstein midpoint from a naive Minkowski average and is what preserves the hyperbolic geometry under aggregation.

\noindent\textbf{Relative geometry encoding.} The relative geometry between region proposals $i$ and $j$ is encoded from centre offsets and log size ratios:
\begin{equation}
\Delta_{ij} = \left[\tfrac{x_i - x_j}{w_j},\, \tfrac{y_i - y_j}{h_j},\, \log\tfrac{w_i}{w_j},\, \log\tfrac{h_i}{h_j}\right] \in \mathbb{R}^4.
\label{supp:eq:spatial}
\end{equation}
This four-dimensional per-pair feature is concatenated with the RoI-feature pair and fed into the spatially-biased multi-head self-attention that produces the dense relation tensor $\mathbf{R}\in\mathbb{R}^{N\times N\times d}$.

\noindent\textbf{RMS text-mean rescaling.} The prefix-token rescaling of the main paper's Eq.~(4) is, in fully explicit form, applied to any projected prefix token $\mathbf{t} \in \mathbb{R}^{d_{\text{llm}}}$ as
\begin{equation}
\hat{\mathbf{t}} = \mathrm{RMSNorm}(\mathbf{t}) \cdot \frac{\|\bar{\mathbf{e}}\|_2}{\|\mathrm{RMSNorm}(\mathbf{t})\|_2},
\label{supp:eq:rmsfix}
\end{equation}
where $\bar{\mathbf{e}}$ is the mean text embedding vector in the batch. The result is a prefix whose L2-norm distribution tracks that of the text embeddings.

\noindent\textbf{Numerical stability notes.} All the primitives above have edge cases that produce NaNs under naive floating-point implementation. We clamp the tangent-vector norm to $[10^{-6}, 5]$ before every $\exp$ map to avoid both underflow in $\sinh(\cdot)/\|\tilde{\mathbf{v}}\|$ and overflow in $\cosh$. The $\arccos$ argument in Eq.~\ref{supp:eq:angle} is clamped to $[-1+10^{-6}, 1-10^{-6}]$ to protect its derivative. The curvature $\kappa$ is clamped to $[0.1, 5.0]$ per gradient step; a hinge penalty $\mathcal{L}_\kappa=\max(0, 0.3-|\kappa|)^2$ discourages $\kappa$ from approaching the lower bound, which would blow up the cone half-apertures.

\begin{figure}[!htb]
    \centering
    \includegraphics[width=.9\linewidth]{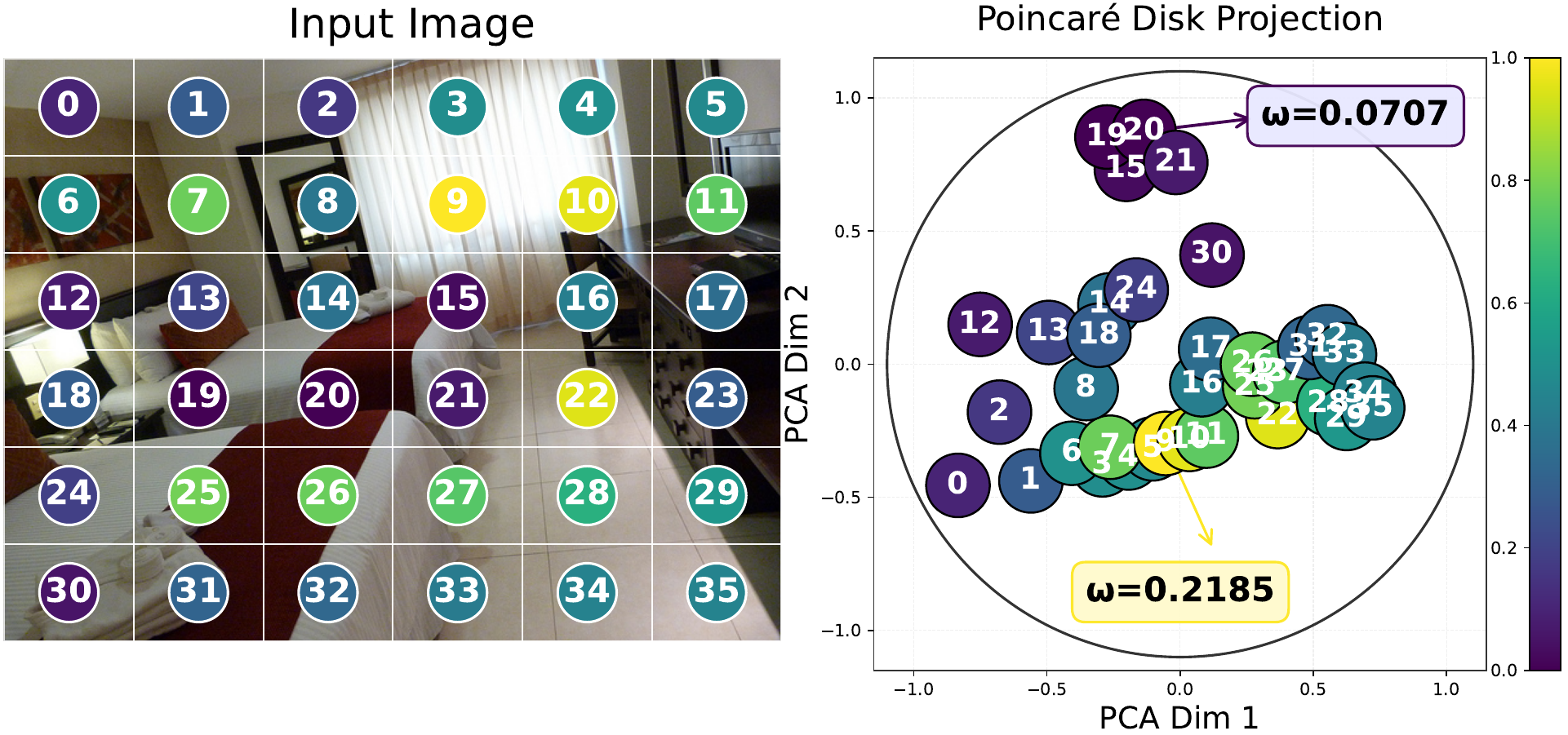}
    \caption{Poincar\'e disk projection of the learned Lorentz manifold. Wider entailment cones near the origin correspond to more general regions (e.g., ``room'', ``field''); specific objects (e.g., ``table'', ``horse'') sit near the boundary, reflecting the IoA-driven spatial hierarchy learnt from containment cues.}
    \label{supp:fig:manifold}
\end{figure}

\noindent Fig.~\ref{supp:fig:manifold} shows the geometry these losses produce: broad scene-level nodes near the origin, specific instance nodes toward the boundary, with cone apertures decreasing with specificity.

\section{Training and Compute Details}
\label{supp:sec:training}

The main paper Sec.~2.3 sketches the training setup in one paragraph. This section gives the full recipe (Table~\ref{supp:tab:hyperparams}) so an implementer can reproduce the reported numbers.

\begin{table}[!htb]
\centering
\caption{Full training recipe for the headline configurations. Values are shared across configurations unless noted. Per-configuration deltas ($\alpha_{\max}$ for the two attention-bias runs, prefix dropout for the prefix runs, LoRA target modules for the capacity control) are stated where the configuration is introduced.}
\label{supp:tab:hyperparams}
\scalebox{0.7}{
\begin{tabular}{ll}
\toprule
\textbf{Group} & \textbf{Setting} \\
\midrule
Backbone & LLaVA-1.5-7B (frozen weights) \\
LoRA rank & 16 (query and value projections; all seven linear projections for the capacity control) \\
Optimizer (LoRA) & AdamW, learning rate $2\times 10^{-5}$, weight decay 0 \\
Optimizer (graph) & Riemannian Adam~\cite{geoopt}, learning rate $10^{-4}$ \\
Optimizer ($\kappa$) & Vanilla Adam, learning rate $10^{-4}$, no gradient clipping \\
\midrule
Batch size & 8 per GPU, 2 GPUs per run, gradient accumulation 2 (16 samples per logged step) \\
Precision & bfloat16 \\
Training data & GQA \texttt{train\_balanced}: 943{,}000 questions over 72{,}140 images (scene-graph boxes for all) \\
Epochs & 5 for S1; 3 for every other configuration (176{,}800 logged steps) \\
Warmup & 100 steps linear ramp on $\alpha(t)$ from $0.001$ to $\alpha_{\max}$ \\
\midrule
Manifold dimension & $n = 256$ \\
Number of region proposals & $N = 36$ (scene-graph boxes at training time; $6{\times}6$ grid at evaluation) \\
RoI feature dimension & $d_v = 1024$ (CLIP ViT-L, before projector) \\
Prefix tokens (Mechanism B) & $K = 4$ \\
Prefix dropout & $p = 0.5$ (S4 sets $p = 0$) \\
Cone half-aperture constant & $K_\omega = 0.1$ (in Eq.~\ref{supp:eq:aperture}) \\
Curvature bounds & $\kappa \in [0.1, 5.0]$, initialised at $1.0$ \\
\midrule
Qwen2-VL-7B / InternVL3-8B & S5 recipe ported; LoRA r16, learning rate $5\times10^{-5}$, 4{,}000 optimiser steps, gate init 0 \\
\bottomrule
\end{tabular}}
\end{table}

\noindent\textbf{Optimizer group separation.} The three optimizer groups matter for stability. Putting the LoRA weights and the hyperbolic-graph parameters under the same optimizer causes the graph to run at the LoRA learning rate ($2\times 10^{-5}$), which is 5$\times$ too small for the geometric losses to converge within the training budget. Putting the curvature $\kappa$ under Riemannian Adam causes it to over-shoot the $[0.1, 5.0]$ interval because the Riemannian correction inflates its step size near the lower bound. Vanilla Adam for $\kappa$ alone, without gradient clipping, was the setting under which $\kappa$ reliably stabilised near its initialised value.

\noindent\textbf{Compute cost.} Every LLaVA configuration was trained on two NVIDIA H100 80GB GPUs. Three-epoch runs take 25--26 hours of wall-clock time measured from checkpoint timestamps (soft prompt 25.4 h, RMS prefix 26.0 h, G2 26.2 h). The Qwen2-VL-7B and InternVL3-8B closure runs (4{,}000 steps) take about two hours each on one H100. Full-testdev GQA evaluation is 45--60 minutes per configuration on a single H100; a single ablation pass over all 12{,}578 questions with two bias settings takes about 30 minutes. VSR and Winoground evaluations take under five minutes each; COCO captioning on val2017 takes about 30 minutes per configuration with greedy decoding.

\noindent\textbf{Ported backbones.} Qwen2-VL-7B pre-expands image tokens and derives 3D M-RoPE positions from the input ids, so the collator reserves $K$ pad-token slots after the image span and the forward pass overwrites their embeddings rather than splicing new tokens; the sequence length is unchanged and the pathway is caption-invariant either way. InternVL3-8B tiles each image dynamically, so training used batch size 1 and RoI features were taken from the global thumbnail tile, the only tile whose frame matches the normalised boxes. Only the S5 recipe (raw gated hyperbolic prefix) was ported.

\noindent\textbf{Reproducibility.} Most LLaVA configurations are single training runs; the RMS prefix (three runs), the soft prompt (two), the Euclidean containment control (three) and LoRA-all-linear (two) have replicates, and the training-seed standard deviation across five matched runs is ${\approx}0.2$pp on relational GQA. Code, checkpoints, the per-item outputs behind every table, and the table-generating script are released at \url{https://github.com/mrfarazi/the-gate-always-closes}.

\section{Behavioural Ablations of the Auxiliary Pathway: All Designs, Full Sweeps}
\label{supp:sec:ablations}

Main paper Table~3 reports the ablation of the five designs at the learned gate. This section gives the same measurement for every gated-prefix checkpoint and the full gate sweep for S5.

\noindent\textbf{What is measured.} Every arm goes through one implementation with $6{\times}6$ grid proposals. For attention-bias checkpoints the $(B,1,P,P)$ patch bias is multiplied by $m\in\{0, 0.1, 0.3, 1, 3\}$ ($m{=}1$ is the learned bias; the main paper reports $m{=}0$, $1$ and $3$, and the intermediate multipliers lie between; for S3 the learned Stage-2 head and layer gates and $\beta{=}{-}8.496$ are applied before $m$). For prefix checkpoints the pre-gate prefix tokens are multiplied by a forced $\gamma$ ($\gamma{=}0$ keeps four zero vectors in the slot; ``learned'' is the trained $\tanh\gamma$; $\gamma{=}1$ is fully open), and \emph{absent} removes the slots entirely. Items are fixed and paired; intervals are 5{,}000-resample paired bootstraps against $\gamma{=}0$ (or $m{=}0$). GQA-rel is the first 1{,}000 testdev questions of semantic type \texttt{rel} in file order; SugarCrepe is the first 1{,}000 pairs, round-robin over its seven sub-tasks.

\begin{table}[!htb]
\centering
\caption{Ablation of every gated-prefix checkpoint (GQA-rel $n{=}1{,}000$; SugarCrepe $n{=}1{,}000$). $\Delta$ is learned $-$ zeroed with paired 95\% CI. Every RMS-normalised design moves by at most 1.5pp at its learned gate; the un-normalised S5 does not. The \emph{absent} column shows the 2--5pp cost of keeping zeroed slots at all.}
\label{supp:tab:gated_prefix}
\scalebox{0.6}{
\begin{tabular}{@{}lr rrrrl rrrrl@{}}
\toprule
 & & \multicolumn{5}{c}{GQA-rel ($n{=}1{,}000$)} & \multicolumn{5}{c}{SugarCrepe ($n{=}1{,}000$)} \\
\cmidrule(lr){3-7}\cmidrule(lr){8-12}
checkpoint & $\tanh\gamma$ & $\gamma{=}0$ & learned & $\gamma{=}1$ & absent & $\Delta$(learned$-$0) [CI] & $\gamma{=}0$ & learned & $\gamma{=}1$ & absent & $\Delta$ [CI] \\
\midrule
base\_r1 (RMS+tanh) & $-0.154$ & 49.9 & 50.2 & 53.3 & 53.2 & $+0.3\ [-0.5, +1.2]$ & 79.4 & 80.7 & 77.5 & 76.9 & $+1.3\ [+0.6, +2.1]$ \\
soft\_prompt & $-0.196$ & 51.6 & 51.1 & 52.8 & 53.7 & $-0.5\ [-1.3, +0.3]$ & 79.1 & 79.3 & 78.2 & 77.3 & $+0.2\ [-0.5, +0.9]$ \\
no\_dropout (S4) & $-0.362$ & 50.5 & 50.6 & 53.1 & 53.4 & $+0.1\ [-1.1, +1.3]$ & 80.1 & 81.5 & 79.5 & 77.1 & $+1.4\ [+0.5, +2.3]$ \\
euclidean & $+0.185$ & 51.0 & 50.4 & 50.7 & 53.5 & $-0.6\ [-1.3, +0.0]$ & 79.4 & 78.9 & 77.8 & 76.9 & $-0.5\ [-1.1, +0.1]$ \\
soft prompt, 2nd run (batch 96) & $+0.119$ & 49.6 & 48.1 & 49.9 & 54.3 & $-1.5\ [-2.6, -0.5]$ & 79.9 & 80.4 & 82.8 & 77.2 & $+0.5\ [+0.0, +1.1]$ \\
lora\_all\_linear r1 & $+0.024$ & 54.5 & 54.2 & 55.1 & 54.9 & $-0.3\ [-0.9, +0.3]$ & 80.6 & 79.8 & 80.6 & 78.8 & $-0.8\ [-1.4, -0.3]$ \\
lora\_all\_linear s2 & $-0.030$ & 54.3 & 54.6 & 53.6 & 54.2 & $+0.3\ [-0.4, +1.0]$ & 80.3 & 80.1 & 78.7 & 77.9 & $-0.2\ [-0.7, +0.3]$ \\
raw\_gated (S5) & $+0.005$ & 50.7 & 44.0 & 12.2 & 54.3 & $-6.7\ [-8.4, -5.0]$ & 78.5 & 80.0 & 79.5 & 76.8 & $+1.5\ [+0.7, +2.4]$ \\
\bottomrule
\end{tabular}
}
\end{table}

\begin{table}[!htb]
\centering
\caption{Full forced-gate sweep for S5 (raw gated prefix, learned $\tanh\gamma{=}{+}0.0052$). The effect is sign-dependent, grows with $|\gamma|$ on GQA-rel, and helps SugarCrepe at small positive $\gamma$: the pathway is used, harmfully for generation and helpfully for log-likelihood scoring.}
\label{supp:tab:s5_sweep}
\scalebox{0.78}{
\begin{tabular}{@{}l rl rl@{}}
\toprule
 & \multicolumn{2}{c}{GQA-rel ($n{=}1{,}000$)} & \multicolumn{2}{c}{SugarCrepe ($n{=}1{,}000$)} \\
\cmidrule(lr){2-3}\cmidrule(lr){4-5}
$\gamma$ & acc & $\Delta$ vs $\gamma{=}0$ [95\% CI] & acc & $\Delta$ [95\% CI] \\
\midrule
$0$ & 50.7 & $+0.0\ [+0.0, +0.0]$ & 78.5 & $+0.0\ [+0.0, +0.0]$ \\
$+0.0052$ (learned) & 44.0 & $-6.7\ [-8.4, -5.0]$ & 80.0 & $+1.5\ [+0.7, +2.4]$ \\
$-0.0052$ ($-$learned) & 53.5 & $+2.8\ [+1.5, +4.2]$ & 78.0 & $-0.5\ [-1.4, +0.4]$ \\
$0.001$ & 50.5 & $-0.2\ [-1.0, +0.6]$ & 79.1 & $+0.6\ [+0.1, +1.2]$ \\
$0.003$ & 47.4 & $-3.3\ [-4.7, -2.0]$ & 79.4 & $+0.9\ [+0.3, +1.6]$ \\
$0.01$ & 36.4 & $-14.3\ [-16.7, -12.0]$ & 81.2 & $+2.7\ [+1.6, +3.9]$ \\
$0.03$ & 39.2 & $-11.5\ [-13.9, -9.2]$ & 85.3 & $+6.8\ [+5.1, +8.5]$ \\
$0.1$ & 33.9 & $-16.8\ [-19.5, -14.2]$ & 84.7 & $+6.2\ [+4.5, +7.9]$ \\
$0.3$ & 49.1 & $-1.6\ [-3.7, +0.5]$ & 82.3 & $+3.8\ [+1.7, +5.9]$ \\
$1$ & 12.2 & $-38.5\ [-41.7, -35.2]$ & 79.5 & $+1.0\ [-1.6, +3.6]$ \\
$-1$ & 46.5 & $-4.2\ [-7.2, -1.4]$ & 84.3 & $+5.8\ [+4.0, +7.7]$ \\
\midrule
prefix absent & 54.3 & $+3.6\ [+1.8, +5.4]$ & 76.8 & $-1.7\ [-2.8, -0.6]$ \\
\bottomrule
\end{tabular}
}
\end{table}

\noindent\textbf{Reading.} With real proposals, the attention-bias designs S1--S3 and every RMS-normalised prefix are behaviourally closed: ablating the content moves accuracy by at most 1.5pp with intervals covering zero, and the bias multiplied by three changes nothing either (Table~3 of the main paper; Table~\ref{supp:tab:gated_prefix}). The two LoRA-all-linear seeds, whose gates ended with opposite signs, are equally inert. S5 is used (Table~\ref{supp:tab:s5_sweep}). Four further observations bear on protocol design. First, four zeroed slots cost 2--5pp of relational GQA relative to no slots for every prefix design trained with LoRA on the query and value projections (the two LoRA-all-linear runs show none), so ``pathway off'' and ``pathway absent'' are different controls and papers should say which they report. Second, the item-level flip rate under real proposals is 2--4\% for the attention-bias designs, 8--10\% for S4 and 14--20\% for S5, against 10--29\% disagreement between genuinely different models. Third, a few RMS-design shifts at the learned gate are small but resolvable (S4 $+1.4$pp $[0.5, 2.3]$ and base $+1.3$pp $[0.6, 2.1]$ on SugarCrepe; synthesis $-1.5$pp $[-2.6, -0.5]$ on GQA-rel), so ``behaviourally closed'' is a statement about magnitude ($\le1.5$pp), not exact invariance. Finally, forcing the RMS content fully on ($\gamma{=}1$) returns GQA-rel to the absent-slot level (base 53.3 vs.\ 53.2 absent; S4 53.1 vs.\ 53.4) and removes the SugarCrepe gain of the zeroed slot: the four slots act as a learned soft prompt whose effect the real content neutralises, and the learned gate keeps each model at the zero-slot point.

\section{Matched Inference Mode: Contrasts, Seeds and Prefix Scale}
\label{supp:sec:matched}

Main paper Table~5 gives dropped and active accuracies per configuration; this section gives the paired contrasts between configurations in dropped mode, the multi-seed VSR replication, and the prefix-scale sweep behind the RMS/None cell of Table~2.

\begin{table}[!htb]
\centering
\caption{Dropped-mode paired contrasts (pathway removed for both members of each pair; 10{,}000-resample paired bootstrap on shared items; two-sided bootstrap $p$). No configuration separates from vanilla on relational or attribute GQA; the flat containment control ties the hyperbolic prefixes; the containment losses add $+0.87$pp (rel) and $+1.37$pp (attr) over the same flat graph trained without them.}
\label{supp:tab:matched_contrasts}
\resizebox{\columnwidth}{!}{
\begin{tabular}{@{}l lr lr lr@{}}
\toprule
 & \multicolumn{2}{c}{GQA-rel ($n{=}5{,}308$)} & \multicolumn{2}{c}{GQA-attr ($n{=}5{,}186$)} & \multicolumn{2}{c}{VSR ($n{=}2{,}113$)} \\
\cmidrule(lr){2-3}\cmidrule(lr){4-5}\cmidrule(lr){6-7}
contrast (pathway dropped) & $\Delta$ [95\% CI] & $p$ & $\Delta$ [95\% CI] & $p$ & $\Delta$ [95\% CI] & $p$ \\
\midrule
RMS r1 $-$ vanilla & $+0.47\ [-0.26, 1.22]$ & 0.227 & $+0.50\ [-0.44, 1.45]$ & 0.305 & $+0.28\ [-0.99, 1.61]$ & 0.700 \\
RMS r2 $-$ vanilla & $+0.34\ [-0.40, 1.07]$ & 0.391 & $+0.12\ [-0.77, 1.02]$ & 0.813 & $+2.32\ [0.99, 3.64]$ & $<$0.001 \\
Euclid.\ containment $-$ vanilla & $+0.24\ [-0.55, 1.04]$ & 0.540 & $+0.60\ [-0.35, 1.54]$ & 0.229 & $-2.41\ [-3.98, -0.80]$ & 0.004 \\
soft prompt $-$ vanilla & $+0.09\ [-0.77, 0.94]$ & 0.839 & $-1.02\ [-2.12, 0.08]$ & 0.070 & $-2.60\ [-4.21, -0.99]$ & 0.002 \\
Euclid.\ gated $-$ vanilla & $-0.62\ [-1.45, 0.23]$ & 0.160 & $-0.77\ [-1.87, 0.35]$ & 0.176 & $-3.36\ [-5.06, -1.70]$ & $<$0.001 \\
G2 $-$ vanilla & $-0.60\ [-1.45, 0.24]$ & 0.181 & $-0.98\ [-2.10, 0.13]$ & 0.090 & $-2.84\ [-4.40, -1.23]$ & $<$0.001 \\
attn.\ bias $-$ vanilla & $-0.51\ [-1.36, 0.36]$ & 0.260 & $+1.35\ [0.46, 2.26]$ & 0.003 & $-11.74\ [-13.91, -9.56]$ & $<$0.001 \\
Euclid.\ containment $-$ Euclid.\ gated (flat: geo.\ loss vs.\ none) & $+0.87\ [0.34, 1.41]$ & 0.001 & $+1.37\ [0.69, 2.04]$ & $<$0.001 & $+0.95\ [0.09, 1.80]$ & 0.034 \\
RMS r2 $-$ Euclid.\ containment (hyp.\ vs.\ flat) & $+0.09\ [-0.60, 0.77]$ & 0.836 & $-0.48\ [-1.45, 0.48]$ & 0.343 & $+4.73\ [2.93, 6.44]$ & $<$0.001 \\
RMS r1 $-$ Euclid.\ containment (hyp.\ vs.\ flat) & $+0.23\ [-0.38, 0.83]$ & 0.492 & $-0.10\ [-0.93, 0.73]$ & 0.844 & $+2.70\ [1.51, 3.88]$ & $<$0.001 \\
RMS r2 $-$ soft prompt & $+0.24\ [-0.47, 0.96]$ & 0.531 & $+1.14\ [0.08, 2.22]$ & 0.034 & $+4.92\ [3.12, 6.67]$ & $<$0.001 \\
\bottomrule
\end{tabular}
}
\end{table}

\begin{table}[!htb]
\centering
\caption{VSR with the pathway dropped, three training seeds per geometry ($n{=}2{,}113$ statements each). The hyperbolic mean exceeds the flat mean and vanilla, but neither seed-level test clears $p{=}0.05$ at three seeds; we report it as a trend, not a result.}
\label{supp:tab:vsr_seeds}
\scalebox{0.85}{
\begin{tabular}{@{}l l r r@{}}
\toprule
group (VSR, pathway dropped, $n{=}2{,}113$) & seed accuracies & mean & sd \\
\midrule
hyperbolic (\texttt{rms\_ungated} r1 / r2 / r3) & 66.49 / 68.53 / 68.53 & 67.85 & 1.17 \\
flat (\texttt{euclidean\_geo} r1 / s2 / s3) & 63.80 / 66.82 / 65.45 & 65.36 & 1.52 \\
vanilla & 66.21 & 66.21 & -- \\
\midrule
contrast & $\Delta$ & $t$ (df) & $p$ \\
\midrule
hyperbolic $-$ flat (Welch) & $+2.49$ & 2.25 (3.8) & 0.092 \\
hyperbolic $-$ vanilla (one-sample) & $+1.64$ & 2.42 (2) & 0.137 \\
flat $-$ vanilla (one-sample) & $-0.85$ & -0.97 (2) & 0.433 \\
\bottomrule
\end{tabular}
}
\end{table}

\begin{table}[!htb]
\centering
\caption{Prefix-scale sweep for the ungated RMS prefix (2{,}000 GQA testdev questions in file order; 2{,}000 SugarCrepe pairs) and the HyperVis raw prefix under the scene-graph path (constant prefix) versus grid proposals. $s{=}1$ is the trained operating point; $s{=}0$ keeps zeroed slots; \emph{absent} removes them. Below $s{\approx}0.3$ the RMS prefix costs nothing on GQA; at $s{=}1$ it costs 12pp and gains 9pp of SugarCrepe.}
\label{supp:tab:prefix_scale}
\scalebox{0.78}{
\begin{tabular}{@{}l rl rl rrr@{}}
\toprule
 & \multicolumn{4}{c}{RMS prefix, ungated (\texttt{rms\_ungated}), $n{=}2{,}000$} & \multicolumn{3}{c}{HyperVis raw prefix, $n{=}2{,}000$} \\
\cmidrule(lr){2-5}\cmidrule(lr){6-8}
scale $s$ & GQA & $\Delta$ [95\% CI] & SC & $\Delta$ [95\% CI] & GQA grid & GQA, no proposals & SC \\
\midrule
$0$ & 60.8 & $+0.0\ [+0.0, +0.0]$ & 75.8 & $+0.0\ [+0.0, +0.0]$ & 59.0 & 59.0 & 73.4 \\
$0.0003$ & 60.8 & $-0.1\ [-0.3, +0.2]$ & 75.9 & $+0.1\ [-0.2, +0.4]$ & 59.1 & 59.0 & 73.6 \\
$0.001$ & 60.7 & $-0.1\ [-0.4, +0.1]$ & 75.9 & $+0.1\ [-0.3, +0.4]$ & 59.0 & 59.1 & 73.4 \\
$0.003$ & 60.8 & $+0.0\ [-0.3, +0.3]$ & 75.6 & $-0.2\ [-0.6, +0.2]$ & 58.9 & 58.8 & 73.3 \\
$0.01$ & 61.0 & $+0.1\ [-0.2, +0.5]$ & 76.0 & $+0.1\ [-0.2, +0.5]$ & 59.2 & 58.9 & 73.9 \\
$0.03$ & 60.7 & $-0.1\ [-0.5, +0.1]$ & 76.1 & $+0.2\ [-0.1, +0.7]$ & 59.1 & 58.4 & 75.5 \\
$0.1$ & 60.8 & $+0.0\ [-0.5, +0.4]$ & 76.7 & $+0.9\ [+0.4, +1.4]$ & 59.0 & 58.9 & 77.8 \\
$0.3$ & 61.5 & $+0.7\ [-0.2, +1.6]$ & 77.5 & $+1.7\ [+1.1, +2.4]$ & 58.9 & 58.8 & 76.5 \\
$1$ & 48.5 & $-12.3\ [-14.1, -10.5]$ & 85.2 & $+9.3\ [+7.8, +10.8]$ & 50.4 & 34.7 & 79.9 \\
$-1$ & 61.2 & $+0.4\ [-0.4, +1.2]$ & 75.3 & $-0.5\ [-1.1, +0.1]$ & 31.4 & 34.8 & 80.3 \\
\midrule
prefix absent & 61.3 & $+0.4\ [-0.9, +1.7]$ & 75.2 & $-0.7\ [-1.5, +0.1]$ & 59.9 & 59.9 & 73.0 \\
\bottomrule
\end{tabular}
}
\end{table}

\section{Gate-Gradient and Attention Probes}
\label{supp:sec:probes}

\begin{table}[!htb]
\centering
\caption{Gate gradient at $\gamma{=}0$ on the RMS-prefix checkpoint, central differences ($\varepsilon{=}0.01$), $n{=}120$ SugarCrepe caption pairs. The discriminative (paired) gradient is not resolvable while the single-caption cross-entropy gradient on the same items is; the curvature is not resolvable.}
\label{supp:tab:gradient}
\scalebox{0.85}{
\begin{tabular}{@{}lrrr@{}}
\toprule
quantity at $\gamma{=}0$ (\texttt{rms\_ungated}, $n{=}120$, $\epsilon{=}0.01$) & mean & s.e. & $|t|$ \\
\midrule
discriminative pair gradient $\partial(\ell^+{-}\ell^-)/\partial\gamma$ (Prop.~1: should be ${\approx}0$) & $+0.674$ & 3.512 & 0.19 \\
single-caption task-CE gradient (positive control) & $-15.894$ & 6.571 & 2.42 \\
second-order (curvature) term & $-18.368$ & 1208.461 & 0.02 \\
\bottomrule
\end{tabular}
}
\end{table}

\begin{table}[!htb]
\centering
\caption{Attention mass on the four injected slots, averaged over answer-side queries and all 32 decoder layers, as a multiple of the chance level $4/L{=}0.0065$ ($n{=}100$ GQA-rel items, grid proposals). ``0'': slots hold exact zero vectors; ``learned'': trained gate; ``1'': content forced fully on. For the seven gated designs trained with LoRA on q/v the slots draw 33--40$\times$ chance both with zero vectors and at the learned gate; forcing the content fully on lowers the mass to ${\sim}2\times$ for the RMS-gated designs, to 10--11$\times$ for the Euclidean and raw prefixes and 26$\times$ for the second soft-prompt run, while the ungated RMS prefix and the LoRA-all-linear runs sit at 12--21$\times$ throughout: the mass is a property of the slot, not of the signal.}
\label{supp:tab:attention}
\scalebox{0.78}{
\begin{tabular}{@{}llrrrrrrr@{}}
\toprule
checkpoint & gate & L0 & L8 & L16 & L24 & L31 & mean & $\times$chance \\
\midrule
\multicolumn{9}{@{}l}{\emph{(a) layer profile at the learned gate} ($n{=}100$, mass in \%)} \\
\texttt{rms\_ungated} & learned & 1.7 & 17.5 & 14.1 & 9.0 & 7.7 & 13.5 & 20.6 \\
\texttt{rms\_ungated} & open & 1.7 & 17.5 & 14.1 & 9.1 & 7.6 & 13.5 & 20.6 \\
\texttt{soft\_prompt} & learned & 1.5 & 28.1 & 26.4 & 19.1 & 4.0 & 24.6 & 37.7 \\
\texttt{soft\_prompt} & open & 1.0 & 1.0 & 1.3 & 1.4 & 2.8 & 1.4 & 2.1 \\
\texttt{raw\_gated} & learned & 1.5 & 27.8 & 25.6 & 18.4 & 4.5 & 24.0 & 36.7 \\
\texttt{raw\_gated} & open & 1.5 & 5.9 & 7.0 & 3.9 & 2.7 & 7.3 & 11.1 \\
\midrule
\multicolumn{9}{@{}l}{\emph{(b) gate conditions per checkpoint} ($n{=}100$, 6$\times$6 grid, mass in \%)} \\
\texttt{base\_r1} & $\gamma{=}0$ & 1.2 & 28.0 & 26.2 & 19.1 & 3.8 & 24.5 & 37.5 \\
 & learned $-0.154$ & 1.4 & 29.6 & 27.9 & 20.8 & 4.4 & 26.1 & 39.9 \\
 & $\gamma{=}1$ & 1.0 & 1.5 & 1.5 & 1.7 & 2.8 & 1.4 & 2.2 \\
\texttt{soft\_prompt} & $\gamma{=}0$ & 1.2 & 27.4 & 25.8 & 18.5 & 3.5 & 24.0 & 36.8 \\
 & learned $-0.196$ & 1.5 & 28.1 & 26.4 & 19.1 & 4.0 & 24.6 & 37.7 \\
 & $\gamma{=}1$ & 1.0 & 1.0 & 1.3 & 1.5 & 2.8 & 1.4 & 2.1 \\
\texttt{no\_dropout} & $\gamma{=}0$ & 1.2 & 25.5 & 23.5 & 16.2 & 4.4 & 21.9 & 33.5 \\
 & learned $-0.362$ & 2.0 & 30.4 & 28.0 & 20.0 & 5.8 & 26.1 & 40.0 \\
 & $\gamma{=}1$ & 0.8 & 1.6 & 0.8 & 0.8 & 1.8 & 1.1 & 1.7 \\
\texttt{euclidean} & $\gamma{=}0$ & 1.2 & 27.2 & 25.3 & 17.9 & 3.6 & 23.6 & 36.1 \\
 & learned $+0.185$ & 0.9 & 27.1 & 25.2 & 17.8 & 3.2 & 23.4 & 35.9 \\
 & $\gamma{=}1$ & 0.8 & 8.5 & 6.9 & 2.5 & 2.9 & 6.3 & 9.6 \\
\texttt{synthesis} & $\gamma{=}0$ & 1.2 & 26.1 & 24.8 & 18.1 & 4.1 & 23.2 & 35.5 \\
 & learned $+0.119$ & 1.3 & 27.0 & 25.5 & 18.8 & 4.1 & 24.0 & 36.7 \\
 & $\gamma{=}1$ & 1.4 & 20.0 & 18.9 & 11.2 & 6.6 & 16.7 & 25.5 \\
\texttt{raw\_gated} & $\gamma{=}0$ & 1.2 & 26.8 & 24.9 & 17.4 & 4.0 & 23.2 & 35.5 \\
 & learned $+0.005$ & 1.5 & 27.8 & 25.6 & 18.4 & 4.5 & 24.0 & 36.7 \\
 & $\gamma{=}1$ & 1.5 & 5.9 & 7.0 & 3.9 & 2.7 & 7.3 & 11.1 \\
\texttt{rms\_ungated} & $\gamma{=}0$ & 1.2 & 11.9 & 8.9 & 4.3 & 2.6 & 8.7 & 13.4 \\
 & learned (ungated = 1) & 1.7 & 17.5 & 14.1 & 9.0 & 7.7 & 13.5 & 20.6 \\
 & $\gamma{=}1$ & 1.7 & 17.5 & 14.1 & 9.0 & 7.7 & 13.5 & 20.6 \\
\texttt{lora\_all\_linear\_r1} & $\gamma{=}0$ & 1.2 & 10.7 & 8.4 & 4.1 & 3.5 & 8.2 & 12.5 \\
 & learned $+0.024$ & 1.2 & 10.5 & 8.1 & 3.9 & 3.6 & 8.1 & 12.4 \\
 & $\gamma{=}1$ & 1.5 & 14.8 & 12.0 & 6.6 & 3.7 & 11.6 & 17.7 \\
\texttt{lora\_all\_linear\_s2} & $\gamma{=}0$ & 1.2 & 14.3 & 11.0 & 5.2 & 3.2 & 10.6 & 16.3 \\
 & learned $-0.030$ & 1.2 & 15.4 & 12.1 & 5.9 & 3.2 & 11.6 & 17.7 \\
 & $\gamma{=}1$ & 1.4 & 1.4 & 0.9 & 0.8 & 1.9 & 1.3 & 2.0 \\
\bottomrule
\end{tabular}
}
\end{table}

\noindent\textbf{Reading the gradient probe.} Lemma~1 of the main paper makes the first-order discriminative gradient an inner product $\langle\Delta(I), g(\mathbf{b}(I))\rangle$; hypotheses H1--H2 say that inner product is not resolvable at the batch sizes used. Table~\ref{supp:tab:gradient} measures it: $|t|{=}0.19$ for the paired objective against $|t|{=}2.42$ for the single-caption cross-entropy gradient obtained with the same apparatus on the same checkpoint. The positive control is what makes the null informative.

\noindent\textbf{Reading the attention probe.} At the learned gate the injected tokens of the RMS prefix draw 13.5\% of answer-side attention (20.6$\times$ chance) at every depth from layer 8 on, which one might read as ``attended everywhere, used nowhere''. Measuring the same quantity with the slots zeroed (Table~\ref{supp:tab:attention}) shows the mass is there without any signal, and for the RMS-gated designs of main paper Table~3 it \emph{falls} to about $2\times$ chance when the content is forced fully on. Attention onto reserved positions is a sink and cannot indicate use, in either direction. The layer profile (near chance at layer 0, a broad peak over layers 8--24, back down at layer 31) is the same in every condition.

\section{Per-Category SugarCrepe Breakdown}
\label{supp:sec:sc_breakdown}

The main paper reports SugarCrepe as a single average. Here we decompose it across the seven categories (Replace-Object/\allowbreak Attribute/\allowbreak Relation, Swap-Object/\allowbreak Attribute, Add-Object/\allowbreak Attribute; Table~\ref{supp:tab:sc_detail}). These evaluations use grid proposals with the pathway on (the RMS prefix at scale~1).

\begin{table}[!htb]
\centering
\caption{SugarCrepe per-category accuracy (full benchmark, 7{,}514 pairs; grid proposals, pathway on). R: Replace, S: Swap, A: Add. G2 = Hyp.\ prefix with the geometric losses removed (cf.\ main paper Table~4). Strict accuracy under log-likelihood scoring; bold marks the per-column best.}
\label{supp:tab:sc_detail}
\scalebox{0.8}{%
\begin{tabular}{lccccccc|c}
    \toprule
    \textbf{Method} & \textbf{R-O} & \textbf{R-A} & \textbf{R-R} & \textbf{S-O} & \textbf{S-A} & \textbf{A-O} & \textbf{A-A} & \textbf{Avg} \\
    \midrule
    Vanilla & 80.21 & 77.16 & 74.96 & 77.64 & 85.29 & 46.99 & 73.55 & 73.69 \\
    Attn.\ bias (small $\alpha$) & 81.11 & 79.31 & 77.31 & 78.86 & 86.04 & 50.73 & 76.73 & 75.73 \\
    \textbf{RMS prefix, ungated (prefix on)} & \textbf{89.10} & \textbf{87.44} & \textbf{85.14} & \textbf{85.37} & \textbf{89.79} & \textbf{69.30} & \textbf{88.15} & \textbf{84.90} \\
    Soft prompt & 81.60 & 80.71 & 77.81 & 80.08 & 86.04 & 53.73 & 79.05 & 77.00 \\
    Hyp.\ prefix, no geo.\ (G2) & 83.17 & 81.73 & 78.88 & 80.89 & 86.94 & 58.20 & 82.08 & 78.84 \\
    \bottomrule
    \end{tabular}}
\end{table}

The breakdown makes two things visible that the average hides. First, the attention-bias variant is not uniformly stronger than vanilla; its gain is concentrated on the Add perturbations. Second, the ungated RMS prefix with its prefix on beats every other row in every category (84.90\% average) while the same setting costs 12pp of GQA (Table~\ref{supp:tab:prefix_scale}), the log-likelihood side of the trade analysed in Sec.~4 of the main paper; among the gated recipes the G2 row tops every column although its training removed the geometry. Both observations are consistent with the main paper's conclusion that the compositional gains of these recipes come from LoRA fine-tuning and prefix effects rather than from geometric content used at inference.

\section{Generation-Side Evaluation: POPE and COCO Captioning}
\label{supp:sec:generation}

Main paper Sec.~5.3 summarises the generation-side evaluation; the full table is below. The RMS-prefix row evaluates that recipe with the prefix zeroed (slots kept); the prefix-dropout row has the pathway absent (LoRA-only); the attention-bias row has its bias on with grid proposals; the soft prompt has its content-free prefix on. A POPE run with the RMS prefix on at scale~1 (grid proposals) is included.

\begin{table}[!htb]
\centering
\caption{Generation-side evaluation. POPE reports accuracy (\%) on the random, popular and adversarial object-existence probes ($3{,}000$ questions each). Captioning is COCO val2017 (5{,}000 images), greedy decoding, pycocoevalcap (SPICE omitted: Java incompatibility in our evaluation environment). Inference modes are stated in the text above.}
\label{supp:tab:generation}
\resizebox{\columnwidth}{!}{
\begin{tabular}{@{}l rrr rrrr@{}}
\toprule
 & \multicolumn{3}{c}{POPE accuracy (\%, $n{=}3{,}000$ each)} & \multicolumn{4}{c}{COCO captioning} \\
\cmidrule(lr){2-4}\cmidrule(lr){5-8}
config & random & popular & adversarial & BLEU-4 & METEOR & ROUGE-L & CIDEr \\
\midrule
LLaVA-1.5 vanilla & 86.30 & 85.13 & 83.40 & 0.324 & 0.280 & 0.562 & 1.132 \\
attn.\ bias (small $\alpha$) & 86.30 & 85.53 & 84.13 & 0.314 & 0.255 & 0.545 & 1.069 \\
RMS prefix, ungated & 83.97 & 83.30 & 82.30 & 0.321 & 0.259 & 0.550 & 1.096 \\
soft prompt & 86.63 & 85.90 & 84.20 & 0.287 & 0.235 & 0.532 & 0.987 \\
RMS prefix + 50\% prefix dropout (\texttt{rms\_dropout\_r1}) & 86.40 & 85.70 & 84.07 & -- & -- & -- & -- \\
RMS prefix, ungated, \emph{prefix on} ($s{=}1$, grid proposals) & 85.50 & 84.80 & 83.37 & -- & -- & -- & -- \\
\bottomrule
\end{tabular}
}
\end{table}

\noindent\textbf{Reading.} With the prefix zeroed, the RMS-prefix recipe is 1--2pp below vanilla on POPE and closest to vanilla on every captioning metric, while the soft prompt is POPE-clean but loses 13\% of CIDEr; the attention-bias model with its bias on matches vanilla on POPE and loses 6\% of CIDEr. Training the RMS recipe with 50\% prefix dropout brings POPE back to the vanilla level (86.40/85.70/84.07) with the pathway absent. With the RMS prefix genuinely on ($s{=}1$), POPE is $85.5/84.8/83.4$, within 0.8pp of vanilla and above the zeroed condition on every split, while the same checkpoint and scale lose 12pp of GQA (Table~\ref{supp:tab:prefix_scale}); yes/no existence questions are not where an injected relational prefix costs. POPE alone therefore underestimates generation harm in both directions: the soft prompt is POPE-clean and captioning-degraded, and the active prefix is POPE-clean and GQA-degraded.

\noindent\textbf{POPE subset construction.} POPE probes ``Is there an X in the image?'' where X is either a positive (present) or a distractor (absent) object. \emph{Random} draws distractors uniformly from the COCO label set, \emph{popular} biases toward objects that frequently co-occur with the present objects, and \emph{adversarial} picks distractors semantically near the present objects; adversarial is the hardest split.

\section{Extended Discussion}
\label{supp:sec:discussion}

This section extends the Conclusion of the main paper with the discussion threads that space did not permit in the main text.

\noindent\textbf{What would a non-suppressed injection look like?}
Closure arises from two interacting causes: (i)~the auxiliary signal is caption-invariant, so the discriminative gradient through the gate factors as in Lemma~1 and is not resolvable under H1--H2; and (ii)~the task loss treats the signal as noise or worse. A pathway that avoids both would need to be \emph{text-coupled}: its contribution to the hidden states should depend on the text being processed, not only on the image. Query-conditioned bilinear interactions between text queries and geometric features are a natural candidate: each preposition (``on'', ``under'', ``beside'') would then obtain a different lens onto the relational structure, and the pathway would carry the first-order gradient that caption-invariant injections lack. LLaMA-Adapter's zero-initialised attention~\cite{gao2023llamaadapter} is the canonical working case of such a pathway.

\noindent\textbf{What the geometric losses do and do not do.}
One might read the geometric losses as a structural regulariser for LoRA. Under matched evaluation they are not, in-distribution: with the pathway dropped, the hyperbolic and flat containment recipes sit within 0.6pp of vanilla on relational GQA and within 1pp on attribute questions, as do the soft prompt and G2 (Table~\ref{supp:tab:matched_contrasts}), and the containment objective adds under 1.4pp over the same graph trained without it. What the losses visibly do is change what the pathway carries when it is on: the RMS geometric prefixes are the configurations most hurt by injection (7--15pp on relational GQA), whereas the content-free soft prompt costs 3pp. Out of distribution, three hyperbolic seeds average 2.5pp above three flat seeds on VSR with the pathway dropped, a trend at $p{\approx}0.09$ that would need six or seven seeds per group to test; we leave it open.

\noindent\textbf{The compositionality--generation tradeoff is partly a norm tradeoff.}
Unnormalised prefix tokens force the LLM to attend to them by virtue of their large magnitude, which benefits log-likelihood scoring but disrupts autoregressive generation. RMS normalization trades part of this forced signal for generation: at scale~0.3 the RMS prefix is free on GQA and worth $+1.7$pp on SugarCrepe; at scale~1 it costs 12pp of GQA and gains 9pp of SugarCrepe (Table~\ref{supp:tab:prefix_scale}). The trade does not disappear with normalisation; it becomes tunable.

\noindent\textbf{Implications beyond this work.}
We conjecture that auxiliary pathways sharing the two properties our analysis relies on, caption-invariance and a task loss that does not reward the pathway's contribution, are susceptible to the same closure, which would include some adapter, side-network and external-memory designs. We have not tested those architectures, and the conjecture should not be read as a result. What we do recommend, on the evidence here, is procedural: a learnable gate's value and the attention mass onto injected positions are not indicators of use; an ablation with real inputs, reported against both a zeroed and an absent control, is.

\noindent\textbf{Connection to the residual-stream literature.}
The residual-stream-dominance argument in the main paper's Sec.~4 rests on a standard reading from mechanistic interpretability: transformer hidden states can be read as a sum of contributions written by earlier blocks and never explicitly subtracted. The 36$\times$ initial-norm gap is a large-perturbation instance of the same phenomenon that mechanistic-\allowbreak interpretability studies observe at smaller scales when they inspect the residual stream at intermediate layers. In our setting the effect is loud enough that a simple RMS rescaling changes the outcome; more subtle norm mismatches, which are common across VLM adapter architectures, may be silent contributors to reported generation degradations.

\section{Limitations}
\label{supp:sec:limitations}

The main paper's Conclusion states the scope in one paragraph and points here.

\noindent\textbf{Training seeds.} Most LLaVA configurations are single training runs. The replicates we have (three RMS-prefix runs, two soft prompts, three Euclidean containment seeds, two LoRA-all-linear seeds, three seeds on each ported backbone) put the training-seed standard deviation at ${\approx}0.2$pp on relational GQA, an order of magnitude below the ablation and matched-mode effects we rely on (2.7--15.5pp), but the 0.8pp flat-containment effect and the VSR trend are within reach of seed noise and are labelled accordingly.

\noindent\textbf{Scope of Proposition~1.} Lemma~1 is exact; Proposition~1 rests on H1--H2, which we measure on one checkpoint and one objective. The second-order regime is characterised by a derived normalisation well and the S3 witness; the measured curvature is uninformative, and a general statement of when an auxiliary perturbation systematically hurts the loss remains open.

\noindent\textbf{Backbone scope.} Closure is shown for five designs on LLaVA-1.5-7B and for the S5 recipe on Qwen2-VL-7B and InternVL3-8B, at 4{,}000 steps on the ported backbones. The behavioural ablations and matched-mode results are LLaVA-only (plus the forced-gate audits on the other two backbones). Larger models and pathways other than prefix tokens and attention biases are untested.

\noindent\textbf{Evaluation protocol.} The ablations use a fixed $6{\times}6$ grid of proposals because the evaluation sets carry no scene graphs; training used real boxes. A detector-based proposal set at evaluation time could move the on-condition numbers, though not the closure of the gates.

\noindent\textbf{Noise control.} G2 removes the geometric losses but keeps the prefix architecture; a random-tensor prefix under the same recipe would test the NEFTune-style hypothesis~\cite{jain2024neftune} that any perturbation of the input embeddings yields the SugarCrepe gain. ArcFace-style angular baselines need class labels that our spatial supervision lacks. Table~\ref{supp:tab:futurework} lists the follow-ups.

\begin{table}[!htb]
\centering
\caption{Follow-up experiments deferred to future work.}
\label{supp:tab:futurework}
\scalebox{0.88}{
\begin{tabular}{p{0.45\linewidth}p{0.48\linewidth}}
\toprule
\textbf{Experiment} & \textbf{What it would establish} \\
\midrule
Six to seven seeds per geometry on VSR (pathway dropped) & Whether the hyperbolic OOD trend ($p{\approx}0.09$ at three seeds) is real \\
Random-tensor prefix under the RMS recipe & Whether generic input noise reproduces the SugarCrepe gain \\
Text-coupled (query-conditioned) injection & Whether breaking caption-invariance keeps the gate open and the pathway used \\
Detector-based proposals at evaluation & Sensitivity of the on-condition numbers to proposal quality \\
Remaining designs on Qwen2-VL / InternVL3; larger backbones & Whether behavioural closure of the normalised designs replicates beyond LLaVA-1.5-7B \\
Non-hyperbolic structured signals (scene graphs, dense predictions) & Whether closure is signal-agnostic \\
\bottomrule
\end{tabular}}
\end{table}

\end{document}